\documentclass[letterpaper]{article}

\usepackage{aaai2027}

\usepackage[hyphens]{url}
\usepackage{natbib}
\usepackage{caption}

\usepackage{amsmath, amssymb, mathtools, amsthm}
\usepackage{amsfonts}

\usepackage{graphicx}
\usepackage{booktabs}
\usepackage{multirow}
\usepackage{placeins}

\usepackage{algorithm}
\usepackage{algpseudocode}

\usepackage{microtype}
\theoremstyle{plain}
\newtheorem{theorem}{Theorem}[section]
\newtheorem{proposition}[theorem]{Proposition}

\theoremstyle{definition}

\newtheorem{assumption}{Assumption}
\theoremstyle{remark}

\usepackage{xspace}
\newcommand{\VAST}{\emph{VAST}\xspace}
\newcommand{\FlexMatch}{\emph{FlexMatch}\xspace}
\newcommand{\FreeMatch}{\emph{FreeMatch}\xspace}
\newcommand{\CoMatch}{\emph{CoMatch}\xspace}
\newcommand{\SimMatch}{\emph{SimMatch}\xspace}
\newcommand{\ExMatch}{\emph{ExMatch}\xspace}
\newcommand{\FixMatch}{\emph{FixMatch}\xspace}
\newcommand{\Laplace}{\emph{Laplace}\xspace}
\newcommand{\Poisson}{\emph{Poisson}\xspace}
\newcommand{\PoissonMBO}{\emph{PoissonMBO}\xspace}
\newcommand{\CS}{\emph{C\&S}\xspace}
\newcommand{\Iscen}{\emph{LP-DSL}\xspace}
\newcommand{\LaplacianShot}{\emph{LaplacianShot}\xspace}
\newcommand{\app}{Supp.\xspace}
\newcommand{\daphna}[1]{}
\newcommand{\itai}[1]{}
\newcommand{\torm}[1]{}

\newif\ifshowmain  \showmaintrue    
\newif\ifshowsupp                   
\showsupptrue                     
\nocopyright
\usepackage{xr}
\ifshowmain
  \title{{Follow the Geometry, Not the Model: Cold Start Semi-Supervised Learning}}
\else
  \title{Supplementary Material for\\ Follow the Geometry, Not the Model: Cold Start Semi-Supervised Learning}
\fi

\author{Itai David and Daphna weinshall}
\affiliations{School of Computer Science and Engineering, The Hebrew University of Jerusalem, Jerusalem 91904, Israel\\ itai.david1@mail.huji.ac.il, daphna@mail.huji.ac.i}

\begin{document}

\maketitle

\ifshowmain

\begin{abstract}
Modern semi-supervised learning (SSL) couples pseudo-label generation and classifier training, using the classifier's own confidence to select the pseudo-labels that are then used to update the model. In the cold-start regime, where at most a few labels per class are available, this coupling is ill-posed, since the classifier cannot supervise itself before it has learned. To address this problem, we propose \VAST (Veracity-Aware Semi-Supervised Training), which decouples these two stages. Probabilistic beliefs over the unlabeled set are first inferred directly from the geometry of a frozen self-supervised embedding and only then distilled into an inductive classifier. The construction rests on the \textbf{Veracity Matrix}, a kernel-based structure that aggregates label evidence across the data manifold and admits an interpretation as a Dirichlet posterior under a per-observation powered-likelihood model. Additionally, we introduce \textbf{Veracity Propagation}, a self-terminating belief-spreading step that extends coverage beyond the kernel neighborhood of the labeled set. Under a controlled protocol in which all methods receive identical frozen embeddings and labeled sets, \VAST outperforms the strongest graph-based SSL baselines at every operating point across three datasets, with statistically significant gains in 7 of 9 comparisons, while producing a deployable inductive classifier rather than requiring transductive graph inference. Compared with end-to-end confidence-gated SSL, we further find that these methods underperform in this setting and, in our experiments, do not consistently exceed labeled-only performance.
\end{abstract}

\section{Introduction}
\label{sec:introduction}

The success of deep learning is often contingent on large, labeled datasets.
In many practical settings - such as medical imaging, satellite imagery, or rare-event detection, obtaining labels is expensive, requiring domain expertise or costly annotation pipelines.
Semi-supervised learning (SSL) addresses this by exploiting the structure of unlabeled data alongside a small labeled set.

State-of-the-art SSL pipelines such as \FixMatch, \FlexMatch, \FreeMatch, \CoMatch\ and \SimMatch\ share a common structure: the classifier's softmax confidence selects the pseudo-labels that are then used to train it further. The construction is sound whenever enough labels exist to calibrate the classifier before pseudo-labeling begins. In the \textbf{cold-start} regime, defined here as at most four labeled samples per class on datasets with many classes, that precondition fails. The model's softmax outputs cannot yet separate informative pseudo-labels from noise, and the training loop either saturates with overconfident but incorrect labels or starves its own confidence gate. This is precisely the regime in which SSL is most valuable, and precisely the regime in which its standard supervision signal is unavailable.

We adopt a different construction. Instead of asking what a trained model predicts for an unlabeled point, we ask what the geometry of the labeled set, viewed through a frozen self-supervised embedding, implies about that point. This separates pseudo-label inference from classifier training: beliefs are formed from geometry alone, before any classifier is fit, and the classifier consumes them afterwards through distillation. The classifier therefore becomes the consumer of pseudo-labels rather than their producer, which removes the circular dependency that makes cold-start SSL ill-posed.

\begin{figure}[t]
    \centering
    \includegraphics[width=\linewidth]{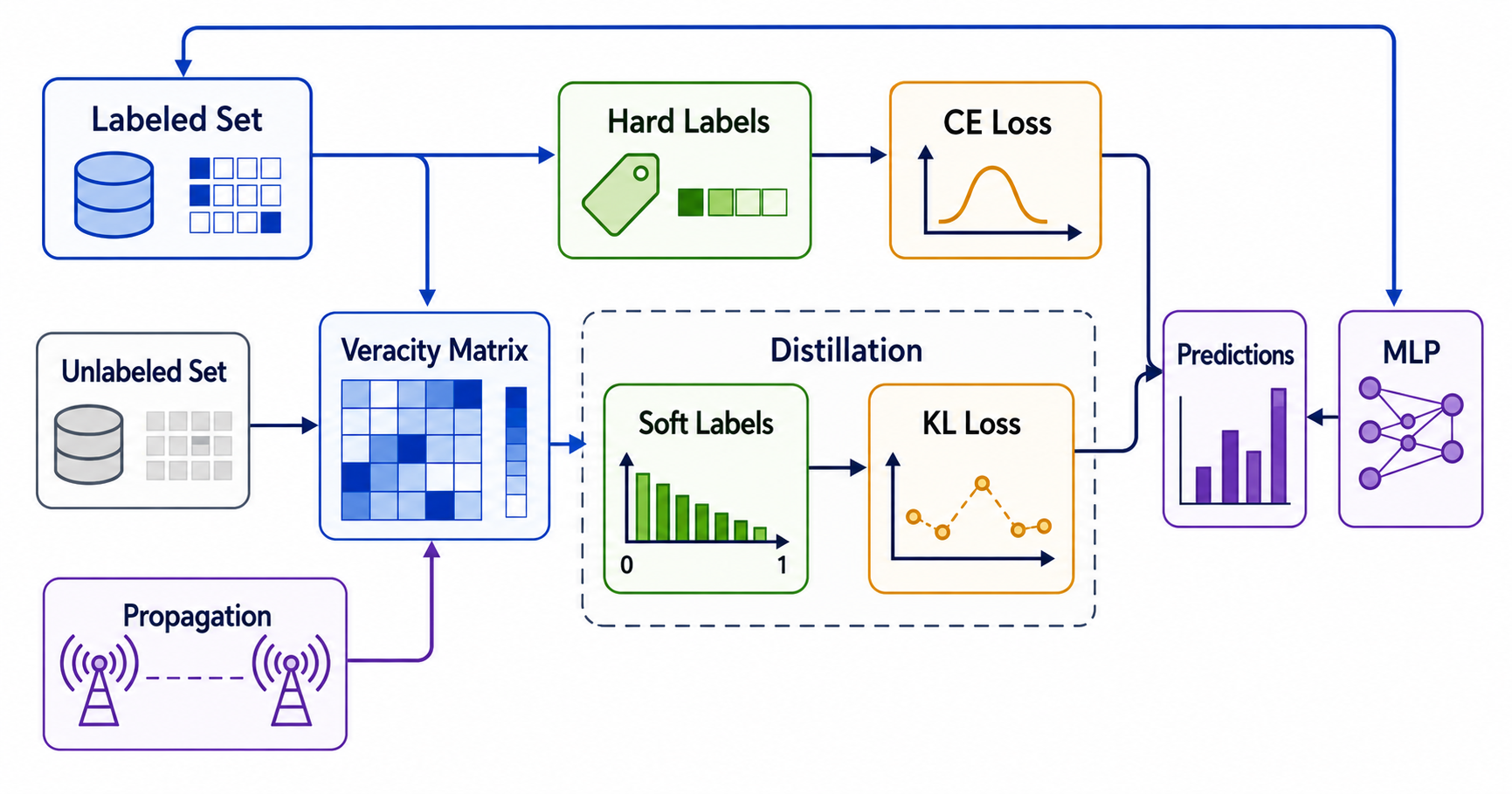}
    \caption{\VAST pipeline: labeled and unlabeled data feed the Veracity Matrix, refined via Propagation, then supervise the classifier through cross-entropy and distillation losses.}
    \label{fig:VAST_pipeline}
    
\end{figure}

\paragraph{Our method}
(Figure~\ref{fig:VAST_pipeline}). Given a small labeled set and a self-supervised embedding, we construct the \textbf{Veracity Matrix} $V \in \mathbb{R}^{N \times M}$, where each row encodes a per-point Dirichlet-Categorical belief over the $M$ class labels,
equivalent to a Dirichlet posterior under a per-observation generalization of the
power-prior framework
(Section~\ref{subsec:veracity_matrix} and \app~\ref{app:bayesian_justification}).
This belief is built by propagating label evidence through an RBF kernel similarity structure, where points that are geometrically close to labeled examples of class $c$ accumulate evidence for class $c$.
Unlabeled points whose veracity belief is sufficiently peaked (high-confidence) are selected as soft pseudo-label targets. The total training loss combines standard cross-entropy on the labeled set with a distillation term on these soft pseudo-labels.

\paragraph{Results.}
Under a controlled protocol in which every method receives identical frozen embeddings and identical labeled sets per seed (Section~\ref{subsec:main_results}, Table~\ref{tab:crossdataset}), \VAST\ improves over the strongest classical graph-SSL baseline at every label budget on all three datasets, with statistically significant gains in 7 of 9 cells, and returns an inductive model where \Poisson\ and \CS\ require a transductive re-solve per query. A complementary comparison against end-to-end confidence-gated SSL on CIFAR-100 (Table~\ref{tab:track2}) places \VAST\ at $39.2\%$ with one label per class, against $30.7\%$ for \CoMatch\ and $9.8\%$ for \FreeMatch, values that reflect the lack of convergence rather than a graded loss of accuracy (Section~\ref{subsec:main_results}).

\paragraph{Our contributions.}
\begin{enumerate}
    \item \textbf{Semi-supervised learning as inference before learning.} We treat pseudo-label inference as a problem in its own right, solved on the geometry of a frozen representation before any classifier exists, so that the classifier consumes supervision instead of producing it.

    \item \textbf{Supervision carried by uncertainty, not by decisions.} Label evidence is kept as a posterior over classes rather than collapsed to an assignment, so one quantity is at once the training target, the weight on that target, and the test for admitting it. Being a property of the geometry rather than of a model, it also lets the method set its own neighborhood size, prior, stopping time and gate without labels, which is crucial in a regime where no validation set is available~\citep{oliver2018realistic}.

    \item \textbf{Controlled evidence and inductive deployment.} On identical frozen embeddings across three datasets, \VAST\ leads every baseline at every budget, while remaining inductive while the transductive baselines require a graph re-solve per query.
\end{enumerate}

\section{Related Work}
\label{sec:related_work}

\subsection{Semi-Supervised Learning}

The dominant paradigm in modern SSL is \textit{pseudo-labeling with consistency regularization}~\citep{laine2017temporal,tarvainen2017mean}, enforcing agreement between a model's predictions under perturbation or across training checkpoints.
\FixMatch~\citep{sohn2020fixmatch} thresholds the model's softmax confidence to generate hard pseudo-labels, enforcing consistency between weakly and strongly-augmented views. \FlexMatch~\citep{zhang2021flexmatch} and \FreeMatch~\citep{wang2023freematch} refine this with class-adaptive and self-adaptive thresholds, respectively, and numerous variants extend the same paradigm~\citep{chen2023softmatch,zheng2022simmatch,li2021comatch,berthelot2019mixmatch,xie2020uda}.
A related line pairs self-supervised pretraining with few-label
distillation (SimCLRv2~\citep{chen2020big}) or non-parametric
support-set soft labels (PAWS ~\citep{assran2021semi}); VAST
differs by deriving supervision from an explicit
kernel-propagated belief over the full unlabeled pool.
We refer to this line of threshold-based methods collectively as the
\mbox{*Match} family, those four members of which (\FlexMatch, \FreeMatch, \CoMatch, \SimMatch) are evaluated as baselines in
Section~\ref{subsec:main_results}.
Oliver et al.~\cite{oliver2018realistic} document evaluation pitfalls in
SSL, including validation-set assumptions incompatible with genuinely
low-label regimes; this motivates our fixed prior $\alpha_0=1/M$
(Section~3.2) and label-free connectivity criterion for $k$ (Section~3.3).

While often achieving state-of-the-art performance, these methods all rely on the classifier's confidence to select pseudo-labels. This assumption is likely to break down in the cold-start regime, where pseudo-label quality degrades sharply~\citep{lucas2021barely}. \ExMatch~\citep{kim2024exmatch} mitigates this effect but still gates pseudo-labels through the model's own predictions. In contrast, \VAST eliminates this dependence entirely: like \FreeMatch, it thresholds a confidence score, but that score is derived from the geometry of a self-supervised embedding rather than the classifier. Consequently, pseudo-label quality is independent of model calibration and remains reliable from the first training step. This motivates the two-track evaluation of Section~\ref{subsec:setup_exp}.

\subsection{Label Propagation and Graph-Based SSL}

Label Propagation (LP) methods propagate labels through a graph defined by feature similarity. Classical formulations~\citep{zhu2003semi,zhou2004learning} solve a global diffusion system over the full graph but degenerate at very low label rates, where solutions are dominated by graph structure and converge toward near-uniform assignments. \Poisson\ Learning~\citep{calder2020poisson} replaces the Laplacian with a Poisson equation, making it the strongest classical baseline in our setting.
Correct and Smooth (\CS)~\citep{huang2021correct} is architecturally close to \VAST: it trains a classifier and then propagates its predictions via residual correction, whereas \VAST\ propagates label information first and trains the classifier afterwards. \citet{iscen2019label} (\Iscen) likewise pair LP with neural training, using hard pseudo-labels from a single LP solve. \LaplacianShot~\citep{ziko2020laplacian}, originally a few-shot method, instead regularizes predictions over the unlabeled batch with a Laplacian term. GNN-based methods such as GCN~\citep{kipf2017semi} and APPNP~\citep{klicpera2019predict} learn propagation and classification jointly and are most naturally compared transductively. We evaluate \Laplace,  \Poisson, \CS\, \LaplacianShot, and \Iscen\ on the same embeddings as \VAST\ (Section~\ref{subsec:main_results}); all first four are transductive, since classifying a new point requires re-solving the system on the enlarged graph, whereas \Iscen\ and \VAST\ are inductive.


\subsection{Self-Supervised Representations}

\VAST operates on frozen self-supervised embeddings rather than raw pixels.
We use BYOL~\citep{grill2020bootstrap} as our primary backbone, since it produces well-structured embeddings which are suitable for kernel-based propagation. Additionally, Barlow Twins~\citep{zbontar2021barlow}, SimCLR~\citep{chen2020simclr} and the externally pretrained DINOv2~\citep{oquab2024dinov2}, are evaluated (Section~\ref{subsec:setup_exp} and \app~\ref{app:embeddings}).

\section{Method}
\label{sec:method}

\subsection{Problem Setup}
\label{subsec:setup}

Let $\mathcal{D} = \{x_i\}_{i=1}^{N}$ denote a dataset of $N$ samples with feature embeddings $\phi(x_i) \in \mathbb{R}^d$ obtained from a pretrained self-supervised encoder $\phi$.
We have a small labeled set $\mathcal{L} = \{(x_k, y_k)\}_{k=1}^{N_L}$ with $N_L \ll N$, and an unlabeled pool $\mathcal{U} = \{x_i\}_{i=1}^{N_U}$, $N_U = N - N_L$.
$M$ denotes the number of classes.

\begin{assumption}[Smoothness]
\label{asm:smoothness}
The embedding $\phi$ is assumed to satisfy the \emph{smoothness constraint}~\citep{chapelle2006semi} when points that are nearby in the embedding space are likely to belong to the same class.
\end{assumption}

\VAST does not commit to a particular representation: 
the Veracity Matrix, the training objective, and the propagation step operate on the embedding as a given input.
Any encoder satisfying Assumption~\ref{asm:smoothness} can be used; we use BYOL as a concrete instantiation and additionally Barlow Twins, SimCLR, and the externally pretrained DINOv2   (Section~\ref{subsec:main_results} and \app~\ref{app:embeddings}).

Next, we define the RBF kernel:
\begin{equation}
    K(x, x') = \exp\!\left(-\frac{\|\phi(x) - \phi(x')\|^2}{2\sigma^2}\right)
    \label{eq:rbf}
\end{equation}
where $\sigma$ is the kernel bandwidth. Embeddings are L2-normalized, bounding pairwise squared distances in $[0,4]$; we fix $2\sigma^2=1$, spanning the full dynamic range of the geometry, and do not sweep it. The kernel is evaluated only on the edges of a sparse neighborhood graph and normalized for local density as described in Section~\ref{subsec:sparsity}.

\subsection{The Veracity Matrix}
\label{subsec:veracity_matrix}

The core of \VAST is the \textbf{Veracity Matrix} $V \in \mathbb{R}^{N \times M}$.
Each entry $V_{k,c}$ accumulates kernel-weighted evidence that point $x_k$ belongs to class $c$, from all labeled points of that class:
\begin{equation}
    V_{k,c} = \alpha_0 + \sum_{(x_j, y_j) \in \mathcal{L}} \mathbb{I}(y_j = c) \cdot K(x_j, x_k)
    \label{eq:unnormalized_veracity}
\end{equation}
where $\alpha_0 > 0$ is a uniform Dirichlet prior.
We interpret this construction as a Bayesian posterior update under a \emph{powered-likelihood} (tempered posterior) model~\citep{ibrahim2015power,friel2008marginal,bhattacharya2019bayesian}, generalized here to per-observation reliability weights rather than a single global discount: each labeled point $(x_j, y_j)$ contributes evidence toward $x_k$'s label with reliability $K(x_j, x_k) \in [0,1]$ rather than as a certain, unit-weight observation, so $V_{k\cdot}$ correspond to the parameters of a Dirichlet posterior $\theta_k \mid \mathcal{L} \sim \mathrm{Dir}(V_{k,1}, \dots, V_{k,M})$ (See~\app~\ref{app:bayesian_justification}). The normalized Veracity vector of point $x_k$ is:
\begin{equation}
    \tilde{V}_{k,c} = \frac{V_{k,c}}{\sum_{m=1}^{M} V_{k,m}}
    \label{eq:normalized_veracity}
\end{equation}
This gives a distribution $\tilde{\mathbf{V}}_k \in \Delta^{M-1}$, which approximates $P(y_k = c \mid \mathcal{L}, x_k)$.
Four properties of this construction matter for what follows: (i) the beliefs are derived purely from labeled-set geometry, with no trained model dependency; (ii) every point aggregates evidence from all labeled points, weighted by manifold proximity; (iii) the peakedness score $\max_c \tilde{V}_{k,c}$ directly measures how unambiguously the local neighborhood assigns a class to $x_k$; and (iv) adding a new labeled point $(x^*, y^*)$ requires only a column update $K(x^*, \cdot)$ to the $y^*$ column of $V$, enabling efficient online operation.
Performance is robust to the choice of peakedness function (Max, Entropy, Margin, unnormalized Margin); we adopt Max by default
(\app~\ref{app:peakedness}).

\paragraph{Sensitivity to $\alpha_0$.}We fix $\alpha_0 = 1/M$ a priori (not tuned per dataset), the natural symmetric-Dirichlet choice with no class-specific bias. Ablation study shows that sweeping $\alpha_0$ over a band around this value moves accuracy by at most $0.53$ points on CIFAR-100 and $0.82$ points on TinyImageNet at any budget, comparable to seed-level noise on both datasets, showing robustness to the choice of $\alpha_0$  (\app~\ref{app:alpha0_ablation}).

\subsection{Graph Sparsification \& Density Normalization}
\label{subsec:sparsity}

Computing the full $N \times N$ kernel matrix is expensive and introduces noise from distant, uninformative pairs.
We therefore evaluate the kernel only on the edges of a sparse neighborhood graph, constructed and normalized as follows.

\paragraph{Connectivity-based $k$-NN graph.}
We build a union-symmetrized $k$-nearest-neighbor graph over the embeddings: an edge $(i,j)$ exists if $x_j$ is among the $k$ nearest neighbors of $x_i$ \emph{or} vice versa.
The neighborhood size $k$ is selected by a purely structural, label-free criterion: \textbf{the smallest $k$ for which the graph forms a single connected component.} This is the minimal connectivity for evidence to reach every point, since smaller $k$ leaves isolated islands and larger $k$ adds distant, increasingly impure edges. The criterion uses no pseudo-labels, no clustering and no tuned threshold, self-adapting to each embedding space: $k=13$ for CIFAR-100 (BYOL), $k=5$ for ImageNet100 (BYOL), $k=3$ for TinyImageNet (SimCLR) (\app~\ref{app:embedding_geometry}).

\paragraph{Density normalization.}
Dense regions still dominate kernel weights on this graph: points in the interior of large clusters accumulate far more (largely redundant) evidence mass than points in sparse regions, which are precisely the points most in need of label information in the cold-start regime~\citep{narayanan2006relation,kim2022defense}; full statistics are in \app~\ref{app:embedding_geometry} and Table~\ref{tab:degree_stats}.
\citet{coifman2006diffusion} address this issue in general form by dividing kernel weights by a power of local degree; we adopt the underlying principle but replace their polynomial penalty with a logarithmic variant tailored to this sparse, hub-prone setting, a substitution also made by \citet{corso2020principal} in the graph-network setting to avoid the over-amplification of linear degree scaling. Using the weighted degree $d_i = \sum_j K(x_i, x_j)$ over graph edges:
\begin{equation}
    \hat{K}(x_i, x_j) = \frac{K(x_i, x_j)}{\sqrt{\log\!\left(1 + d_i\, d_j\right)}}
    \label{eq:degree_norm}
\end{equation}

The logarithm dampens hubs without collapsing their edges, and the square root splits the correction symmetrically across an edge's two endpoints, so $\hat{K}(x_i,x_j) = \hat{K}(x_j,x_i)$. The correction is two-sided by construction: it damps edges where $d_i d_j > e-1$, while lifting weak edges in sparse regions toward unit weight. This ensures that points least able to accumulate initial evidence can effectively overcome the cold-start problem (full derivation in \app~\ref{app:embedding_geometry}). The normalized kernel $\hat{K}$ replaces $K$ throughout; its empirical contribution is quantified in Section~\ref{subsec:ablations}.

\subsection{Veracity Propagation}
\label{subsec:propagation}

Seeding the Veracity Matrix from $\mathcal{L}$ alone leaves any unlabeled point outside the kernel neighborhood of every labeled point with the uniform prior $\alpha_0$; we call these points \textbf{uninformed}. Veracity Propagation extends coverage to this subset by using already-informed unlabeled points as secondary evidence sources, iterating until belief refinement stalls.

At each round $t$, every point is assigned a role based on its peakedness $c_i = \max_c \tilde{V}_{i,c}$:
\begin{itemize}
    \item \textbf{Senders} $\mathcal{S}_t = \{x_i \mid c_i > 1/M\}$: above chance, so they contribute evidence.
    \item \textbf{Receivers} $\mathcal{R}_t = \{x_i \in \mathcal{U} \mid c_i < \tau_r\}$, $\tau_r{=}0.75$: not yet confident; points above $\tau_r$ are frozen to avoid diluting settled beliefs.
\end{itemize}
The roles overlap: a point with $1/M < c_i < \tau_r$ both sends and receives, since partially informed points are the conduits by which evidence reaches uninformed ones.

For each receiver $x_k \in \mathcal{R}_t$:
\begin{equation}
    V_{k,c} \leftarrow V_{k,c} + \sum_{x_s \in \mathcal{S}_t} V_{s,c} \cdot \hat{K}(x_s, x_k)
    \label{eq:propagation}
\end{equation}
Senders contribute their unnormalized accumulated evidence ${V}_k$, not the normalized $\tilde{V}_k$, so a sender's contribution scales with its total accumulated evidence. No self-loops. This is followed by re-normalization to obtain $\tilde{V}_k$ and updated $c_i$ for the next round's role assignment.

\paragraph{Stopping criterion.}
Roles are recomputed after each round. We stop propagation when no new receiver freezes, or equivalently the first round with $\mathcal{F}_{t+1} = \mathcal{F}_t$, where $\mathcal{F}_t = \mathcal{L} \cup \{x_i \in \mathcal{U} : c_i^{(t)} \geq \tau_r\}$ is the frozen set. This replaces both a hand-tuned iteration count and a decay schedule. Because frozen points are never revisited (Eq.~\ref{eq:propagation}), $\mathcal{F}_t$ is monotone non-decreasing, and the following guarantee holds:
\begin{proposition}[Finite termination]
\label{prop:finite_termination}
Let $T^\star$ be the smallest index for which $\mathcal{F}_{T^\star+1} = \mathcal{F}_{T^\star}$. Then $T^\star \leq N - |\mathcal{F}_0| \leq |\mathcal{U}|$, ensuring that propagation halts in at most $|\mathcal{U}|+1$ rounds.
\end{proposition}
\noindent The full proof is given in \app~\ref{app:termination}.
Empirically, termination is fast (10-30 rounds), with smaller-$k$ datasets requiring more since evidence traverses longer paths. See \app~\ref{app:prop_convergence}.

\paragraph{Sensitivity to $\tau_r$.}
The receiver threshold is the only fixed hyperparameter in the propagation stage; \VAST's accuracy is stable across a wide sweep of $\tau_r$ values, with no setting standing out as an outlier (Figure~\ref{fig:tau_r_sensitivity}; full details in \app ~\ref {app:receiver_ablation}).
We use $\tau_r = 0.75$ throughout. 
After propagation terminates, the updated Veracity Matrix is used directly in the \VAST training objective with the distillation gate (Section~\ref{subsec:training}) applied uniformly to all unlabeled points.

\begin{figure}[t]
    \centering
    \includegraphics[width=0.8\linewidth]{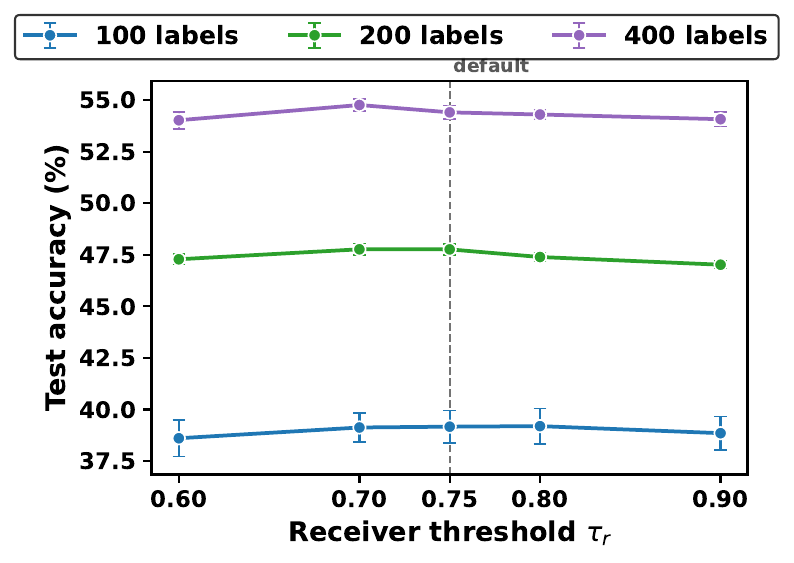}
    \caption{Receiver-threshold ($\tau_r$) sensitivity on CIFAR-100 (BYOL): test accuracy (mean $\pm$ SEM, 5 seeds) at 100/200/400 labels. Dashed line marks the default $\tau_r = 0.75$.}
    \label{fig:tau_r_sensitivity}
\end{figure}

\subsection{Pseudo-Label Generation and Distillation }
\label{subsec:training}

Once the Veracity Matrix has been constructed and enriched via propagation (Section~\ref{subsec:propagation}), it is used to generate soft pseudo-labels for the unlabeled set and to train a classifier via knowledge distillation.

\paragraph{Adaptive, class-balanced gate.}
The gate is set from the shape of the peakedness distribution, not hand-tuned. After propagation, this distribution is bimodal on every dataset we evaluate, and the base threshold $\tau$ is placed in the valley between the two modes (full details and unimodal fallback case in  \app~\ref{app:mode_threshold}).
Poorly separated classes accumulate less evidence and are therefore less peaked, so a single global cutoff would skew $\mathcal{T}_{\text{train}}$ toward the classes the embedding already handles best. We therefore rescale it per class,
\begin{equation}
    \tau_c = \max\!\left(\bar{\mu}_c\,\tau,\; 1/M\right)
    \label{eq:class_threshold}
\end{equation}
with $\bar{\mu}_c \in (0,1]$ the mean belief mass of class $c$ normalized by its largest value over classes, floored at the above-chance bar of Section~\ref{subsec:propagation} ($1/M$).

\paragraph{Pseudo-label preparation.}
For each point $x_i$ in the dataset, we compute its normalized veracity vector $\tilde{\mathbf{V}}_i$ (Eq.~\ref{eq:normalized_veracity}) and its peakedness score $c_i = \max_c \tilde{V}_{i,c}$.
We define a binary distillation mask:
\begin{equation}
    D_i = \mathbb{I}\!\left[c_i > \tau_c\right] \cdot \mathbb{I}\!\left[x_i \in \mathcal{U}\right]
    \label{eq:distillation_mask}
\end{equation}
Points with $D_i = 1$ are admitted into the training set as soft-label targets; points with low peakedness are excluded, since their veracity distribution is considered too diffuse to provide reliable supervision.
The effective training set is now:
\begin{equation}
    \mathcal{T}_{\text{train}} = \mathcal{L} \;\cup\; \{x_i \in \mathcal{U} \mid D_i = 1\}
    \label{eq:training_set}
\end{equation}

\paragraph{Why soft labels, not hard pseudo-labels.}
We use the full normalized veracity vector $\tilde{\mathbf{V}}_i$ as the target rather than its argmax, which preserves uncertainty: a point near a class boundary retains mass on both classes and regularizes the model more gently than a possibly incorrect hard label. This matters most in the cold-start regime, where even peaked assignments carry residual uncertainty.

\paragraph{Training objective.}

The classifier is trained by minimizing a combined loss over mini-batches $\mathcal{B} \subset \mathcal{T}_{\text{train}}$, split into labeled samples $\mathcal{B}_\mathcal{L}$ and distillation samples $\mathcal{B}_\mathcal{U}$:
\begin{equation}
    \mathcal{L}_{\text{total}} = \mathcal{L}_{\text{CE}} +  \lambda  \ \mathcal{L}_{\text{distill}}
    \label{eq:total_loss}
\end{equation}
where $\lambda$ controls the relative weight of the distillation loss.
We set $\lambda = 64$ for CIFAR-100 and ImageNet100, and $\lambda = 32$ for TinyImageNet. A sensitivity analysis over $\lambda \in [1, 256]$ on CIFAR-100 shows accuracy varies by less than one point across this range at every label budget (\app~\ref{app:lambda_ablation}). The supervised term is the standard cross-entropy:
\begin{equation*}
    \mathcal{L}_{\text{CE}} = \frac{1}{|\mathcal{B}_\mathcal{L}|}\sum_{(x,y) \in \mathcal{B}_\mathcal{L}} \mathrm{CrossEntropy}(\hat{y}(x),\, y)
\end{equation*}
The distillation term minimizes the KL-divergence between the classifier's output distribution and the veracity belief, weighted per-sample by each point's own peakedness:
\begin{equation*}
\mathcal{L}_{\text{distill}} =  \frac{1}{|\mathcal{B}_\mathcal{U}|}\sum_{\!\!\!(x_i,\,\tilde{\mathbf{V}}_i) \in \mathcal{B}_\mathcal{U}\!\!\!\!\!\!\!\!\!} c_i \cdot \mathrm{KL}\!\left(\mathrm{softmax}(\hat{y}(x_i)) \,\Big\|\, \tilde{\mathbf{V}}_i\right)
\end{equation*}
Above, $c_i = \max_c \tilde{V}_{i,c}$ measures how peaked the label distribution is for point $x_i$ (Section~\ref{subsec:veracity_matrix}).
The normalized veracity vector $\tilde{\mathbf{V}}_i$ is used directly as the target distribution. 
Per-sample weighting and uniform weighting ($w_i=1$) are statistically indistinguishable on CIFAR-100 (\app~\ref{app:distill_weight_ablation}); we adopt the per-sample form as the default.

Algorithm~\ref{alg:VAST} summarizes the complete procedure, from graph construction through classifier training.

{

\begin{algorithm}[t]
\caption{\VAST Training}
\label{alg:VAST}
\begin{algorithmic}[1]
\footnotesize 
\Require Labeled set $\mathcal{L}$, unlabeled pool $\mathcal{U}$, embedding $\phi$, Dirichlet prior $\alpha_0$, distillation weight $\lambda$
\Ensure Trained classifier $f_\theta$
\State $\tau_r  \gets 0.75$ \Comment{receiver threshold} 
\State $k \leftarrow$ smallest $k$ giving a connected union $k$-NN graph over $\phi(\mathcal{L} \cup \mathcal{U})$; build sparse kernel $\hat{K}$ on its edges \Comment{Section~\ref{subsec:sparsity}}
\State $V_{k,c} \leftarrow \alpha_0$ for all $x_k \in \mathcal{L} \cup \mathcal{U}$, $c \in \{1,\dots,M\}$
\State $V_{k,y_j} \leftarrow V_{k,y_j} + \hat{K}(x_j, x_k)$ for each $(x_j, y_j) \in \mathcal{L}$, $x_k$ with $\hat{K}(x_j, x_k) > 0$ \Comment{Veracity Matrix, Section~\ref{subsec:veracity_matrix}}
\State Normalize $\tilde{\mathbf{V}}_k \leftarrow V_{k,\cdot} / \sum_c V_{k,c}$; $\;c_i \leftarrow \max_c \tilde{V}_{i,c}$
\Repeat \Comment{Veracity Propagation, Section~\ref{subsec:propagation}}
    \State $\mathcal{S} \leftarrow \{x_i \mid c_i > 1/M\}$, $\;\mathcal{R} \leftarrow \{x_i \in \mathcal{U} \mid c_i < \tau_r\}$
    \State $V_{k,c} \leftarrow V_{k,c} + \sum_{x_s \in \mathcal{S}} V_{s,c} \cdot \hat{K}(x_s, x_k)$ for all $c$, for each $x_k \in \mathcal{R}$
\State Re-normalize $\tilde{\mathbf{V}}_k \leftarrow V_{k,\cdot}/\sum_c V_{k,c}$ and recompute $c_k$ for each updated row $x_k \in \mathcal{R}$
\Until{$|\mathcal{R}|$ unchanged}
\State $\mathcal{T}_{\text{train}} \leftarrow \mathcal{L} \cup \mathrm{AdaptiveGate}(\{c_i \mid x_i \in \mathcal{U}\})$ 
\While{not converged}
    \State Sample batch $\mathcal{B} \subset \mathcal{T}_{\text{train}}$; split into $\mathcal{B}_\mathcal{L}$, $\mathcal{B}_\mathcal{U}$
    \State $\theta \leftarrow \theta - \eta \nabla_\theta \left[\mathcal{L}_{\text{CE}}(\mathcal{B}_\mathcal{L}) + \lambda \cdot \mathcal{L}_{\text{distill}}(\mathcal{B}_\mathcal{U}; \text{per-sample } c_i)\right]$ 
\EndWhile
\State \Return $f_\theta$
\end{algorithmic}
\end{algorithm}}
\section{Experiments}
\label{sec:experiments}

\subsection{Methodology}
\label{subsec:setup_exp}

\begin{table*}[t]
{\centering
\small
\setlength{\tabcolsep}{3pt}
\begin{tabular}{@{}llllllllll@{}}
\toprule
& \multicolumn{3}{c}{CIFAR-100} & \multicolumn{3}{c}{TinyImageNet} & \multicolumn{3}{c}{ImageNet100} \\
\cmidrule(lr){2-4} \cmidrule(lr){5-7} \cmidrule(lr){8-10}
Method & 1/cls & 2/cls & 4/cls & 1/cls & 2/cls & 4/cls & 1/cls & 2/cls & 4/cls \\
\midrule
MLP \scriptsize{(labeled only)} & $29.4_{\pm0.5}$ & $38.8_{\pm0.5}$ & $47.6_{\pm0.5}$ & $12.76_{\pm0.46}$ & $17.89_{\pm0.89}$ & $23.30_{\pm0.96}$ & $19.7_{\pm1.1}$ & $30.8_{\pm1.1}$ & $38.4_{\pm0.5}$ \\
1-NN & $31.0_{\pm0.6}$ & $37.4_{\pm0.4}$ & $43.4_{\pm0.6}$ & $14.59_{\pm0.75}$ & $18.22_{\pm0.70}$ & $21.58_{\pm0.69}$ & $26.7_{\pm2.2}$ & $33.2_{\pm1.5}$ & $37.3_{\pm0.5}$ \\
\Laplace             & $29.4_{\pm0.9}$ & $40.2_{\pm0.6}$ & $47.7_{\pm0.3}$ & $11.18_{\pm1.65}$ & $17.44_{\pm1.13}$ & $22.82_{\pm0.94}$ & $18.0_{\pm1.7}$ & $31.2_{\pm0.3}$ & $39.0_{\pm0.6}$ \\
\Poisson              & $\mathbf{38.6_{\pm1.0}}$ & $45.1_{\pm0.7}$ & $50.1_{\pm0.4}$ & $16.60_{\pm0.68}$ & $21.06_{\pm0.68}$ & $24.36_{\pm0.51}$ & $31.7_{\pm0.9}$ & $38.5_{\pm0.8}$ & $42.0_{\pm0.6}$ \\
\CS                 & $33.6_{\pm0.5}$ & $44.4_{\pm0.5}$ & $51.2_{\pm0.3}$ & $14.51_{\pm0.91}$ & $20.81_{\pm0.57}$ & $\mathbf{26.89_{\pm1.03}}$ & $28.3_{\pm0.8}$  & $37.8_{\pm0.5}$ & $44.8_{\pm0.5}$ \\
\Iscen & $37.4_{\pm1.0}$ & $44.9_{\pm0.4}$ & $51.1_{\pm0.5}$ & $13.13_{\pm1.14}$ & $17.21_{\pm0.89}$ & $20.15_{\pm0.87}$ & $30.4_{\pm1.0}$ & $39.6_{\pm0.7}$ & $44.1_{\pm0.5}$ \\
\LaplacianShot & $31.1_{\pm0.5}$ & $40.9_{\pm0.5}$ & $48.6_{\pm0.3}$ & $14.60_{\pm0.30}$ & $20.24_{\pm0.27}$ & $25.37_{\pm0.50}$ & $26.7_{\pm0.8}$ & $35.7_{\pm0.2}$ & $41.9_{\pm0.5}$ \\
\VAST (Ours) & $\mathbf{39.2_{\pm0.9}}$ & $\mathbf{47.8_{\pm0.3}}$ & $\mathbf{54.4_{\pm0.4}}$ & $\mathbf{17.71_{\pm1.26}}$ & $\mathbf{22.83_{\pm1.40}}$ & $\mathbf{27.55_{\pm1.61}}$ & $\mathbf{34.9_{\pm1.1}}$ & $\mathbf{43.1_{\pm0.9}}$ & $\mathbf{49.3_{\pm0.2}}$ \\
\midrule
\FreeMatch        & $9.8_{\pm0.3}$ & $43.4_{\pm0.9}$ & $50.8_{\pm1.0}^\dagger$ & $4.80_{\pm0.30}$ & $11.40_{\pm0.20}$ & $20.4$ & $6.55_{\pm0.25}$ & $14.50_{\pm2.20}$ & $45.2$ \\
\bottomrule
\end{tabular}
\caption{Cross-dataset summary: test accuracy (\%) by method and label budget. CIFAR-100/ImageNet100 use BYOL embeddings; TinyImageNet uses SimCLR embeddings.
\footnotesize $^\dagger$ taken directly from the official USB benchmark~\citep{usb2022}.}
The bottom row does not use frozen embeddings; it is reproduced from Table~\ref{tab:track2} for scale reference.
\label{tab:crossdataset}}

\end{table*}

\paragraph{Data and label budgets.}
We evaluate on CIFAR-100~\citep{krizhevsky2009learning}, TinyImageNet~\citep{le2015tiny} and ImageNet100~\citep{tian2020contrastive,deng2009imagenet} at 1, 2 and 4 labels per class (100/200/400 total labels; 200/400/800 for TinyImageNet), targeting the cold-start regime. Labeled sets are drawn uniformly at random subject to exact class balance, and within each (dataset, seed, budget) the same labeled set is shared across all methods.

\paragraph{Embeddings.}
The primary encoder is BYOL for CIFAR-100 and ImageNet100 and SimCLR for TinyImageNet, whose BYOL features are markedly less class-separable (\app~\ref{app:embedding_geometry}); Barlow Twins, SimCLR and DINOv2 are additionally compared on CIFAR-100 (\app~\ref{app:embeddings}). All are trained on the target dataset's own unlabeled split except DINOv2~\citep{oquab2024dinov2}, which is used off the shelf after pretraining on 142M external images (LVD-142M), so its absolute accuracies are not comparable to the in-domain columns. VAE2~\citep{calder2020poisson} serves only as a baseline-attribution check (Section~\ref{subsec:ablations}).

\paragraph{Baselines.}
Track~1 compares \VAST\ against five graph-based SSL methods (\Laplace, \Poisson, \CS, \Iscen and \LaplacianShot; Section~\ref{sec:related_work}) on the same frozen embeddings, so that within-column differences are attributable to the method rather than the representation. Track~2 compares against end-to-end pipelines that learn their own features (\FlexMatch, \FreeMatch, \CoMatch, \SimMatch and \ExMatch), together with Conf-ST, a confidence-thresholded self-training control on \VAST's embeddings.

\paragraph{Implementation.}
Classifier architecture, optimizer, hyperparameters, per-baseline graph settings and the full reproducibility protocol are given in \app~\ref{app:implementation}. Results are averaged over 5/6/5 seeds on CIFAR-100/ImageNet100/TinyImageNet for \VAST\ and the frozen-embedding baselines; significance is assessed with two-sided paired $t$-tests on shared labeled sets.

\subsection{Main Results}
\label{subsec:main_results}

Table~\ref{tab:crossdataset} reports \textbf{Track~1} and Table~\ref{tab:track2} \textbf{Track~2}; alternative-encoder results are given in \app~\ref{app:embeddings} (Table~\ref{tab:embedding_sweep}).

\paragraph{Track 1: Controlled comparison on frozen embeddings.}
Baseline provenance and parameter count are detailed in \app~\ref{app:crossdataset_protocol}, ~\ref{app:params}). Three findings stand out:

(1) \textbf{\VAST leads across the board.} \VAST is the top performer in all
9 cells, significantly so in 7 (\app~\ref{app:track1_significance});
the exceptions are ties with \Poisson (CIFAR-100, 1 label/class, $p{=}0.36$)
and \CS (TinyImageNet, 4 labels/class, $p{=}0.41$). Both are transductive,
requiring a re-solve over the enlarged graph to classify a new point, while
\VAST is inductive (Section~\ref{sec:related_work}): where accuracy ties,
deployment cost does not, as the timing analysis below quantifies.

(2) \textbf{The gain is algorithmic, not representational.} All Track~1 methods
share identical frozen embeddings, so the comparison isolates the algorithm.
\VAST significantly leads \Iscen, its closest architectural match, at every budget on all datasets, differing chiefly in \Iscen's hard,
single-solve pseudo-labels versus \VAST's soft, gated beliefs refined over
rounds. \VAST also beats a 1-NN control everywhere (all $p{<}0.001$), ruling
out nearest-neighbor retrieval.

(3) \textbf{Propagation matters.} Ablating Veracity
Propagation costs several points at every CIFAR-100 budget, even with the
distillation-only variant at its own optimum
(Section~\ref{subsec:analysis_backbone}, Table~\ref{tab:prop_ablation}). Since
that variant keeps soft, gated targets, the loss isolates iterative refinement,
not soft targets alone, as the mechanism behind~(2).

\paragraph{Track 2: Confidence-based pseudo-labeling analysis.}
Table~\ref{tab:track2} contrasts \VAST\ with end-to-end confidence-gated SSL on CIFAR-100, first in its end-to-end form: \FlexMatch, \FreeMatch, \CoMatch, \SimMatch, and \ExMatch, the last of these tailored to the low-label setting. Reproduced baselines use the official USB benchmark~\citep{usb2022}; \ExMatch\ has no public implementation, and its numbers are quoted from the original paper, not independently verified (Table~\ref{tab:track2}).

The failure is visible in the results themselves: \FreeMatch\ reaches $9.8\%$ where an MLP trained on the same 100 labels alone reaches $29.4\%$ (Table~\ref{tab:crossdataset}), so its unsupervised signal is actively harmful rather than merely weak. Per-iteration logs show two distinct mechanisms rather than one-graded failure (\app~\ref{app:match_collapse}): FreeMatch's confidence gate
often saturates on incorrect pseudo-labels within a few hundred
iterations and stays there, while CoMatch's gate closes to zero
almost immediately, and it survives only through an ungated
contrastive graph-regularization term. Consistent with this reading, the margin generally contracts as labels accumulate and calibration becomes attainable, falling from $8.5$ points at 100 labels to $0.9$ at 200 and $3.2$ at 400 against the strongest reproduced \mbox{*Match} entry ($6.7$, $0.9$, $3.2$ if \ExMatch\ is admitted).

These baselines also learn their own representation, so the gap cannot be attributed to confidence itself. Conf-ST isolates it, replacing \VAST's veracity beliefs with the classifier's own softmax under identical embeddings, classifier, and labeled sets, averaged over the same 5 seeds as \VAST (\app~\ref{app:conf_st}). Even with $\tau$ tuned per budget in its favor it trails \VAST\ by $8.8$, $7.9$ and $6.2$ points, and \FreeMatch's self-adaptive gate falls below labeled-only training at 1 label/class. The belief source, not the backbone, accounts for the deficit. Counts of training iterations and backbone parameters for all \mbox{*Match} architectures are in \app~\ref{app:track2_protocol} and~\ref {app:params} respectively.

\begin{table}[t]
\centering
\small
\begin{tabular}{@{}lccc@{}}
\toprule
Method & 100 & 200 & 400 \\
\midrule
\FlexMatch (WRN-28-2)    & 16.2 & 33.4 & -- \\
\FlexMatch (WRN-28-8)    & 9.4  & 40.8 &  $49.85_{\pm1.5}^\dagger$ \\
\FreeMatch (WRN-28-8)    & $9.8_{\pm0.3}$ & $43.4_{\pm0.9}$ &  $50.8_{\pm1.0}^\dagger$ \\
\CoMatch                 & 30.7 & 38.9 &  $39.02_{\pm0.77}^\dagger$ \\
\SimMatch                & 17.2 & 46.9 & $51.18_{\pm1.07}^\dagger$ \\
\ExMatch$^*$              & $32.49_{\pm1.3}$ & $43.71_{\pm0.7}$ & $51.13_{\pm1.7}$ \\
\midrule
\multicolumn{4}{@{}l}{\emph{Frozen features, identical to \VAST}} \\
Conf-ST \scriptsize{(fixed $\tau^\ddagger$)} & $30.4_{\pm0.3}$ & $39.9_{\pm1.0}$ & $48.2_{\pm0.7}$ \\
Conf-ST \scriptsize{(adaptive $\tau$)}       & $27.2_{\pm0.2}$ & $37.1_{\pm0.6}$ & $46.6_{\pm0.7}$ \\
\midrule
\textbf{\VAST (BYOL)}   & $\mathbf{39.2_{\pm0.9}}$ & $\mathbf{47.8_{\pm0.3}}$ & $\mathbf{54.4_{\pm0.4}}$ \\
\bottomrule
\end{tabular}
\caption{Confidence-based pseudo-labeling vs.\ \VAST, CIFAR-100 cold-start regime.
\footnotesize $^*$\ExMatch\ results are taken from the original paper.
$^\dagger$ taken directly from the official USB benchmark~\citep{usb2022}.
$^\ddagger$ $\tau$ selected per budget to maximize Conf-ST accuracy.}

\label{tab:track2}

\end{table}

\paragraph{Computational cost.}
Because \VAST\ operates on frozen features and fits only a lightweight classification head, the full pipeline, including one-time BYOL pretraining, is roughly $20\times$ cheaper than a single \FreeMatch\ run on CIFAR-100 and $4\times$ cheaper on ImageNet100. Within Track~1's shared-embedding protocol it is also the fastest label-exploitation method, by up to $15\times$ over the transductive graph solvers (\app~\ref{app:complexity},~\ref{app:timing}). These numbers indicate that in this regime additional compute does not recover the failure modes above, as, for example, \FreeMatch\ consumes $261.1$ GPU-hours to reach $9.8\%$ at 100 CIFAR-100 labels.

\begin{figure}[t]
\centering
\includegraphics[width=0.75\linewidth]{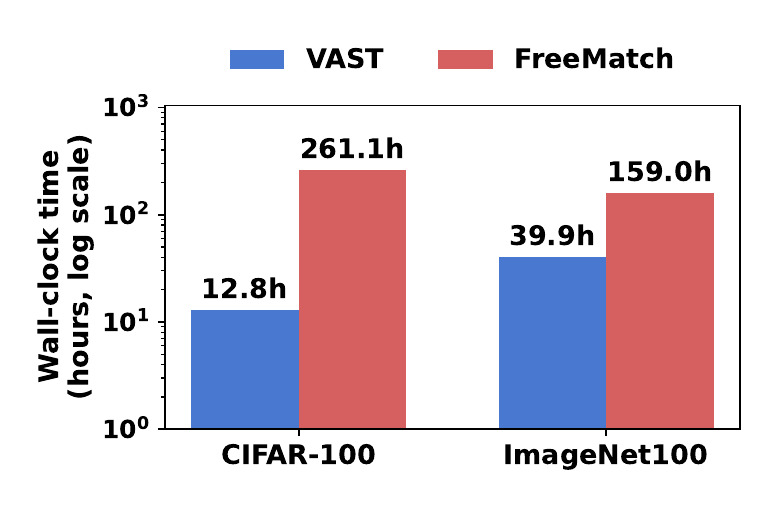}
\caption{Full-pipeline Wall-clock cost, \VAST vs.\ \FreeMatch (log scale, hours). \VAST includes one-time BYOL pretraining; \FreeMatch pays full training cost per run.}
\label{fig:track2_timing}
\end{figure}

\subsection{Ablation Study}
\label{subsec:ablations}

\paragraph{Propagation.}
We ablate the propagation stage (Section~\ref{subsec:propagation}) by training with distillation only, in two variants: \emph{matched kernel} (isolates propagation's contribution) and \emph{density-normalized RBF, best-effort} (re-tuned to compensate for its absence). In both, a constant $\tau$ replaces the mode-based adaptive heuristic, which is calibrated to the post-propagation peakedness distribution and degrades once propagation is removed. Accordingly, propagation is found to be a substantial contributor at every CIFAR-100 budget (Table~\ref{tab:prop_ablation}), replicating on ImageNet100 with comparable margins (\app~\ref{app:prop_ablation_full}).
\begin{table}[t]
\begin{center}
\small
\setlength{\tabcolsep}{4pt}
\begin{tabular}{@{}lccc@{}}
\toprule
Configuration & 1/cls & 2/cls & 4/cls \\
\midrule
\textbf{\VAST (distill + prop.)} & $\mathbf{39.2_{\pm0.9}}$ & $\mathbf{47.8_{\pm0.3}}$ & $\mathbf{54.4_{\pm0.4}}$ \\
Distill only (matched kernel) & $30.5_{\pm0.7}$ & $38.9_{\pm0.6}$ & $47.3_{\pm0.7}$ \\
Distill only (local RBF, best) & $34.9_{\pm1.1}$ & $42.0_{\pm0.9}$ & $48.8_{\pm0.3}$ \\
\bottomrule
\end{tabular}
\caption{Propagation ablation on CIFAR-100: test accuracy (\%) by label budget (labels per class).}\label{tab:prop_ablation}

\end{center}
\end{table}

\paragraph{Degree normalization.}
We ablate the density normalization of the sparse kernel (Section~\ref{subsec:sparsity}, Eq.~\ref{eq:degree_norm}) by removing the log-degree correction and using the raw kernel $K$ directly in Veracity Matrix. This is a strict single-variable swap: $\alpha_0$, $\tau_r$, and the distillation threshold are held at their full-\VAST values. Degree normalization gives a small, consistent accuracy gain at every CIFAR-100 budget (Table~\ref{tab:degree_ablation}), sign-consistent across all 5 seeds and significant at 1 and 4 labels/class and just short at 2 ($p{=}0.06$) (\app~\ref{app:degree_ablation_significance})
\begin{table}[t]
\centering
\small
\setlength{\tabcolsep}{4pt}
\begin{tabular}{@{}lccc@{}}
\toprule
Configuration & 1/cls & 2/cls & 4/cls \\
\midrule
\textbf{\VAST (distill + prop.)} & $\mathbf{39.2_{\pm0.9}}$ & $\mathbf{47.8_{\pm0.3}}$ & $\mathbf{54.4_{\pm0.4}}$ \\
\VAST w/o degree normal. & $37.8_{\pm0.7}$ & $\mathbf{46.8_{\pm0.2}}$ & $53.7_{\pm0.4}$ \\
\bottomrule
\end{tabular}
\caption{Degree-normalization ablation on CIFAR-100: test accuracy by label budget.}
\label{tab:degree_ablation}
\end{table}

\subsection{The Role of the Embedding}
\label{subsec:analysis_backbone}

\VAST\ ranks first on all four encoders at both budgets (Table~\ref{tab:embedding_sweep}), but its advantage is distributed differently as representation quality rises. DINOv2 anchors the high-quality end of this axis, being externally pretrained (LVD-142M) rather than on the target dataset's unlabeled split like the other three. Against a 1-NN classifier on the same features, graph-based inference retains its full value regardless of encoder strength ($+8.2$ points on BYOL, $+11.7$ on DINOv2 at one label per class). What contracts is the difference among solvers: \VAST's margin over the strongest baseline falls from $+2.1$ points on SimCLR to $+0.9$ on Barlow Twins, $+0.6$ on BYOL and $+0.3$ on DINOv2. Cleaner neighbourhoods leave less for any propagation rule to recover, so the solvers converge toward a common ceiling well above trivial transfer.
\VAST\ is therefore distinguished here by robustness rather than peak margin: it alone ranks first on every encoder, while \Laplace\ and \Iscen\ fall below 1-NN on DINOv2 at 1 label/class ($57.1$, $46.5$ vs. $58.4$), recovering only at 4 (\app~\ref{app:embeddings})


\begin{table}[t]
\centering
\small
\setlength{\tabcolsep}{4pt}
\begin{tabular}{@{}lccccc@{}}
\toprule

Method & SimCLR & Barlow & BYOL & DINOv2 & VAE2 \\
\midrule
\multicolumn{6}{@{}c@{}}{\textit{1 label per class}} \\
1-NN      & $24.3_{\pm0.5}$ & $27.64_{\pm0.8}$ & $31.0_{\pm0.6}$ & $58.4_{\pm0.6}$ & -- \\
\Poisson  & $26.83_{\pm0.5}$ & $36.37_{\pm0.8}$ & $\mathbf{38.6_{\pm1.0}}$ & $\mathbf{69.8_{\pm0.9}}$ & $4.5_{\pm0.2}$ \\
\CS       & $23.70_{\pm0.46}$ & $32.76_{\pm0.66}$ & $33.6_{\pm0.5}$ & $67.6_{\pm0.8}$ & -- \\
\VAST     & $\mathbf{28.9_{\pm0.4}}$ & $\mathbf{37.3_{\pm0.5}}$ & $\mathbf{39.2_{\pm0.9}}$ & $\mathbf{70.1_{\pm0.7}}$ & -- \\
\midrule
\multicolumn{6}{@{}c@{}}{\textit{4 labels per class}} \\
1-NN      & $33.29_{\pm0.4}$ & $40.5_{\pm0.5}$ & $43.4_{\pm0.6}$ & $72.56_{\pm0.5}$ & -- \\
\Poisson  & $36.17_{\pm0.5}$ & $48.49_{\pm0.1}$ & $50.1_{\pm0.4}$ & $79.7_{\pm0.3}$ & $7.1_{\pm0.2}$ \\
\CS       & $38.45_{\pm0.36}$ & $49.78_{\pm0.44}$ & $51.2_{\pm0.3}$ & $\mathbf{82.1_{\pm0.2}}$ & -- \\
\VAST     & $\mathbf{41.3_{\pm0.4}}$ & $\mathbf{51.8_{\pm0.4}}$ & $\mathbf{54.4_{\pm0.4}}$ & $\mathbf{82.8_{\pm0.5}}$ & -- \\
\bottomrule
\end{tabular}
\caption{Solver against embedding on CIFAR-100; mean $\pm$ SEM over 5 seeds (\VAST\, \CS\ and 1-NN not run on VAE2). \VAST\ is highest in every column; bold marks a significant margin over the strongest baseline ($p<0.05$, paired $t$-test).}
\label{tab:embedding_sweep}
\end{table}

\section{Limitations}
\label{sec:limitations}

The quality of the frozen embedding bounds VAST's accuracy. The smoothness assumption is the operative requirement, and embeddings that violate it degrade the construction uniformly (Table~\ref{tab:embedding_sweep}). The adaptive distillation gate assumes a bimodal distribution. On low-purity embeddings, the per-class fallback takes over, though sufficiently poor purity would degrade both branches. We assume class-balanced and fixed-class labeled sets. The latter extends, since a new class contributes a column to $V$ without a graph rebuild, but neither setting is evaluated here. Constructing the Veracity Matrix requires the full unlabeled pool in advance, which leaves streaming extensions to future work.
We restrict this study to the cold-start regime with at most 4 labels/class, where the calibration failure characterized applies (Section~\ref{subsec:main_results}); as the margin contraction reported there indicates, we expect confidence-gated methods to become comparable at higher budgets.

\section{Conclusion}
\label{sec:conclusion}

We presented \VAST, a semi-supervised learning method that separates pseudo-label inference from classifier training. Geometry-aware probabilistic beliefs are constructed through the Veracity Matrix and Veracity Propagation, then distilled into an inductive classifier under the smoothness assumption alone. On identical frozen embeddings across three datasets, \VAST\ improves over the strongest classical graph-SSL baseline at every operating point, and remains inductive where most baselines require a transductive re-solve. Against end-to-end confidence-gated SSL, the \mbox{*Match} family does not degrade gracefully in this regime but fails to converge, consistent with the calibration failure that defines it. Future work includes learned kernels that would weaken the smoothness requirement, streaming formulations that remove the up-front unlabeled-pool assumption, and the combination of geometric belief inference with augmentation-based consistency regularization at budgets where the classifier becomes reliable.

\subsection*{\textbf{Acknowledgments}}
This work was supported by a grant from the Gatsby Charitable Foundation and AFOSR award FA8655-24-1-7006.


\bibliography{example_paper}

\fi

\ifshowsupp

\ifshowmain
\section*{Appendix}
\else

This document provides supplementary technical material for the main paper,
organized into four sections. \app~\ref{app:theory} gives the Bayesian
justification of the Veracity Matrix and the finite-termination proof.
\app~\ref{app:algorithms} details the adaptive distillation-threshold
heuristic, its per-class coverage fallback, and the class-balanced threshold
rescaling. \app~\ref{app:protocol} covers implementation and reproducibility
details, full protocol details for the cross-dataset and Track~2 tables, and
backbone parameter counts. \app~\ref{app:additional_results} reports
additional experimental results: significance testing and the \PoissonMBO
comparison; propagation termination; sensitivity to the receiver threshold,
Dirichlet prior, and distillation weight; embedding geometry and the
TinyImageNet BYOL-vs-SimCLR comparison; \mbox{*Match} collapse diagnostics
and the Conf-ST control; complexity and timing; the propagation,
degree-normalization, and distillation-weighting ablations; the peakedness
function choice; and additional embedding results on Barlow Twins, SimCLR,
VAE2, and DINOv2.
References to sections, tables, equations, and assumptions ``of the main
paper'' point to the main submission; all other cross-references are internal
to this document.

\renewcommand{\thetable}{S\arabic{table}}
\renewcommand{\thefigure}{S\arabic{figure}}
\setcounter{table}{0}
\setcounter{figure}{0}
\fi


\appendix

\section{Theoretical Justification}
\label{app:theory}


\subsection{Veracity Matrix and Powered Likelihood}
\label{app:bayesian_justification}

We show that Eq.~2 of the main paper is equivalent to a Dirichlet posterior under a per-observation generalization of the powered-likelihood (tempered posterior) framework~\citep{ibrahim2015power,friel2008marginal,bhattacharya2019bayesian}. This generalization — replacing a single global likelihood discount with a per-pair kernel weight — preserves Dirichlet conjugacy, as shown below, but is not itself a result established in the cited papers; we present it as the natural probabilistic reading of Eq.~2 rather than as a derivation from prior work.

Assume the label of point $x_k$ is sampled from a categorical distribution $\theta_k = (\theta_{k,1}, \dots, \theta_{k,M})$, $\theta_{k,c} \geq 0$, $\sum_c \theta_{k,c} = 1$, with a symmetric Dirichlet prior $\theta_k \sim \mathrm{Dir}(\alpha_0, \dots, \alpha_0)$.
Because the Dirichlet is conjugate to the categorical distribution, observing $n_c$ true labels of class $c$ at $x_k$ would yield posterior $\theta_k \mid \{\text{observations}\} \sim \mathrm{Dir}(\alpha_0 + n_1, \dots, \alpha_0 + n_M)$.

This standard update assumes every observation is a certain, unit-weight sample of the true label.
In our setting, however, a labeled point $x_j$ is not itself an observation \emph{at} $x_k$: rather, it is evidence about $x_k$'s label whose reliability decays with the two points' dissimilarity, as captured by the kernel $K(x_j, x_k) \in [0,1]$.
Plugging in unit counts regardless of proximity would treat every labeled point within kernel reach as an equally certain observation of $x_k$'s label, which is not the intended semantics.

We instead adopt the \emph{tempered posterior} (powered-likelihood) Bayesian framework~\citep{ibrahim2015power,friel2008marginal,bhattacharya2019bayesian}, in which each observation's contribution to the likelihood is raised to a power equal to its reliability before the posterior update.
Under this model, an observation of class $c$ at $x_j$ with reliability $K(x_j, x_k)$ contributes a \emph{discounted soft count} of $K(x_j, x_k)$ rather than $1$ toward class $c$ at $x_k$:
$$
\tilde{n}_{k,c} = \sum_{(x_j,y_j)\in\mathcal{L}} \mathbb{I}(y_j=c)\cdot K(x_j,x_k).
$$
The resulting tempered posterior is again Dirichlet,
$$
\theta_k \mid \mathcal{L} \sim \mathrm{Dir}(\alpha_0+\tilde n_{k,1}, \dots, \alpha_0+\tilde n_{k,M}),
$$
so that $V_{k,c} = \alpha_0 + \tilde n_{k,c}$ (Eq.~2 of the main paper) is exactly the $c$-th Dirichlet parameter of this posterior, and the normalized Veracity vector $\tilde{V}_k$ (Eq.~3 of the main paper) is the posterior mean $\mathbb{E}[\theta_k \mid \mathcal{L}]$.
Under this interpretation, the peakedness score $\max_c \tilde{V}_{k,c}$ can be read as an approximate measure of posterior confidence in $x_k$'s label, rather than as a purely ad hoc heuristic. 
The derivation above applies to the raw RBF kernel $K$, which satisfies $K(x_j,x_k)\in[0,1]$ by construction; the density-normalized variant $\hat{K}$ (Section~\ref{subsec:sparsity} of the main paper) is applied during graph construction and can marginally exceed $1$ in sparse regions, so the Dirichlet posterior reading is an approximation when $\hat{K}$ is used operationally.


\subsection{Finite Termination of Veracity Propagation}
\label{app:termination}

We give the full formal argument behind Proposition~\ref{prop:finite_termination} of the main paper (Section~3.4).

\paragraph{Setup.}
Let $\mathcal{X} = \mathcal{L} \cup \mathcal{U}$ with $|\mathcal{X}| = N$, and let $\tau_r \in (1/M, 1)$ be the receiver threshold. Denote the \emph{normalized} belief matrix at round $t$ by $\tilde V^{(t)} \in \Delta_M$ (Eq.~\ref{eq:normalized_veracity} of the main paper), where $\Delta_M$ is the $M$-simplex, and write $c_i^{(t)} = \max_c \tilde V^{(t)}_{i,c}$. Define the frozen and receiver sets
\begin{align*}
\mathcal{F}_t &= \mathcal{L} \;\cup\; \{i \in \mathcal{U} : c_i^{(t)} \geq \tau_r\}, \\
\mathcal{R}_t &= \mathcal{X} \setminus \mathcal{F}_t.
\end{align*}
Labeled points belong to the frozen set by construction: $\mathcal{L} \subseteq \mathcal{F}_t$ for all $t$ by the definition above, and in particular $\mathcal{L} \subseteq \mathcal{F}_0$. Write $V^{(t)}$ for the unnormalized accumulated-evidence matrix the propagation update (Eq.~\ref{eq:propagation}) actually maintains, so that a sender's contribution scales with its total accumulated evidence rather than only its peakedness (Section~3.4 of the main paper); $\tilde V^{(t)}$ is its row-normalized projection onto $\Delta_M$, recomputed each round for role assignment. In this notation, one round of Eq.~\ref{eq:propagation} is, for all $i \in \mathcal{X}$,
{\scriptsize
$$
\tilde V^{(t+1)}_i \;=\;
\begin{cases}
\tilde V^{(t)}_i & i \in \mathcal{F}_t,\\[3pt]
\mathrm{normalize}\!\left(V^{(t)}_i + \displaystyle\sum_{j \in \mathcal{N}(i)} \hat{K}(x_j, x_i)\, V^{(t)}_j\right) & i \in \mathcal{R}_t,
\end{cases}
\eqno{(A.1)}
$$}
where $\mathcal{N}(i) = \{j \in k\text{-NN}(i) : c_j^{(t)} > 1/M\}$ is the set of neighbors of $i$ acting as senders at round $t$ (any neighbor whose peakedness exceeds chance, following Section~3.4 of the main paper; this set can include neighbors that are themselves still receivers, i.e., $c_j^{(t)} < \tau_r$), and $\hat K$ is the density-normalized kernel of Section~3.3 of the main paper. A frozen row's unnormalized evidence is also left untouched, so its normalized projection is unchanged too, which is the only property the argument below uses: frozen rows are never modified, whether tracked as $V^{(t)}$ or $\tilde V^{(t)}$.

\paragraph{Lemma A.1 (Monotone freezing).}
\emph{For all $t \geq 0$, $\mathcal{F}_t \subseteq \mathcal{F}_{t+1}$.}

\paragraph{Proof.}
Fix $i \in \mathcal{F}_t$. By~(A.1), $\tilde V^{(t+1)}_i = \tilde V^{(t)}_i$, so $c_i^{(t+1)} = c_i^{(t)} \geq \tau_r$. Thus $i \in \mathcal{F}_{t+1}$. \hfill$\blacksquare$

\paragraph{Theorem A.2 (Finite termination).}
\emph{Let $T^\star$ be the smallest index for which $\mathcal{F}_{T^\star+1} = \mathcal{F}_{T^\star}$. Then}
$$
T^\star \;\leq\; N - |\mathcal{F}_0| \;\leq\; |\mathcal{U}|.
$$
\emph{Consequently, the stopping rule ``halt when no receiver freezes'' terminates in at most $|\mathcal{U}| + 1$ rounds.}

\paragraph{Proof.}
By Lemma~A.1, $(\mathcal{F}_t)_{t \geq 0}$ is a non-decreasing sequence of subsets of the finite set $\mathcal{X}$. Whenever $\mathcal{F}_{t+1} \neq \mathcal{F}_t$, monotonicity forces $|\mathcal{F}_{t+1}| \geq |\mathcal{F}_t| + 1$, so $|\mathcal{F}_t|$ can strictly increase at most $N - |\mathcal{F}_0|$ times before saturating. Hence $\mathcal{F}_{T^\star} = \mathcal{F}_{T^\star+1}$ for some $T^\star \leq N - |\mathcal{F}_0| \leq |\mathcal{U}|$, since $\mathcal{L} \subseteq \mathcal{F}_0$. \hfill$\blacksquare$

\paragraph{Assumptions used.}
The proof requires only the finiteness of $\mathcal{U}$ and the monotone-freezing property of update~(A.1). It is independent of kernel properties (positivity, symmetry, connectivity), the normalization form, $\tau_r$'s value beyond $\tau_r > 1/M$, and spectral properties of the graph. In particular, the bound holds even though the sender set $\{j : c_j^{(t)} > 1/M\}$ overlaps with the receiver set.

\paragraph{Scope of the guarantee.}
Theorem~A.2 guarantees termination only. It does \emph{not} imply convergence of $\tilde V^{(t)}$, energy minimization, uniqueness, or contraction, each of which would require a fixed sender set and a spectral bound on the induced operator, neither of which Veracity Propagation provides. The training pipeline (Section~3.5 of the main paper) consumes the stopping rule, not the analytic limit of the belief trajectory; residual-receiver beliefs are handled by the peakedness gate (see below).

\paragraph{Residual receivers.}
Let $\mathcal{R}^\star = \mathcal{R}_{T^\star}$ denote receivers that never cross $\tau_r$. The peakedness gate of Section~3.5 of the main paper excludes them from the distillation set whenever $\min_c \tau_c \geq \tau_r$; even when this does not hold, Eq.~\ref{eq:class_threshold} of the main paper floors every class threshold at $1/M$, so only strictly above-chance beliefs can pass. Membership in $\mathcal{R}^\star$ is itself diagnostic: these are points whose embedding-space evidence remains genuinely ambiguous after propagation.


\section{Algorithmic Details}
\label{app:algorithms}


\subsection{Adaptive Distillation-Threshold Heuristic}
\label{app:mode_threshold}

Algorithm~\ref{alg:mode_threshold} gives the full mode-based threshold
selection procedure used to set the distillation threshold $\tau$
(Section~3.5 of the main paper). The procedure operates on the peakedness
scores of the unlabeled set. It bins the scores into a histogram, smooths it
with a short moving-average kernel, and identifies local maxima (including the
boundary bins, so that spikes at the extremes of the score range count as
modes). Peaks are ranked by height; a secondary peak is accepted as a genuine
mode only if it is separated from the primary peak by at least a fixed
fraction of the histogram width \emph{and} reaches a minimum height relative
to the primary peak; both conditions guard against noise bumps in a decaying
tail being mistaken for real structure. If two genuine modes exist, the
threshold is placed at the left edge of the right-hand mode's bin, its lower
boundary, that is, the valley between the two modes, admitting the
high-confidence population; if only one dominant mode exists, the threshold is
placed at that mode's right edge, though this value is never used, since the
unimodal case instead takes the receiver-threshold $\tau_r$ fallback described
in the Unimodal Fallback discussion below.

\begin{algorithm}[t]
\caption{Mode-Based Threshold Selection}
\label{alg:mode_threshold}
\begin{algorithmic}[1]
\Require Scores $v \in \mathbb{R}_{\geq 0}$; bins $B = 100$; smoothing window $w = 3$; min.\ peak separation $s = 0.05 B$; min.\ relative peak height $h = 0.15$
\Ensure Threshold $\tau$; bimodal indicator $b$
\State $n \leftarrow |v|$
\If{$n = 0$} \Return $(0, \mathrm{false})$ \EndIf
\If{$\max(v) = \min(v)$} \Return $(\min(v), \mathrm{false})$ \EndIf
\State $(H, E) \leftarrow \mathrm{histogram}(v, B)$ over $[\min(v), \max(v)]$ \Comment{$H$: counts, $E$: bin edges}
\State $\bar{H} \leftarrow$ moving average of $H$ with window $w$
\State $P \leftarrow$ indices of local maxima of $\bar{H}$, boundary bins included
\State $p_1 \leftarrow \operatorname*{argmax}_{p \in P} \bar{H}[p]$ \Comment{primary mode}
\State $p_2 \leftarrow$ tallest $p \in P \setminus \{p_1\}$ with $|p - p_1| \geq s$ and $\bar{H}[p] \geq h \cdot \bar{H}[p_1]$, if any
\If{$p_2$ exists}
    \State $r \leftarrow \max(p_1, p_2)$ \Comment{right-hand mode}
    \State \Return $(E[\max(r - 1,\, 0) + 1],\ \mathrm{true})$ \Comment{valley: left edge of the right-hand mode's bin}
\Else
    \State \Return $(E[p_1 + 1],\ \mathrm{false})$ \Comment{right edge of the single dominant mode's bin; superseded by the $\tau_r$ fallback in Algorithm~\ref{alg:coverage_fallback}}
\EndIf
\end{algorithmic}
\end{algorithm}

All four constants ($B$, $w$, $s$, $h$) are fixed structural parameters of the
heuristic, not tuned per dataset; the same values are used in every
experiment.


\subsubsection{Unimodal Fallback}
\label{app:coverage_fallback}

Mode-based thresholding (Algorithm~\ref{alg:mode_threshold}) assumes the
peakedness distribution separates into a low-confidence bulk and a
high-confidence mode. This assumption can fail on sufficiently low-purity
embeddings: the post-propagation peakedness distribution can be unimodal
instead, in which case there is no valley to place $\tau$ in, since the
distribution's own shape no longer supplies a natural base threshold.
Algorithm~\ref{alg:mode_threshold} reports this directly, returning
$b=\mathrm{false}$ when no secondary peak is found, and
Algorithm~\ref{alg:coverage_fallback} falls back to the receiver threshold
$\tau_r$ (Section~3.4 of the main paper) as the base threshold in that case,
then applies the same class-balanced rescaling used in the bimodal branch
(\app~\ref{app:class_balanced_threshold}); this receiver-threshold fallback
with per-class rescaling is the mechanism referred to as the ``per-class
fallback'' in Section~5 of the main paper. This reuses an already-fixed
hyperparameter rather than introducing a second threshold-selection
mechanism, at the cost of no longer adapting the base threshold to the
unimodal distribution's own shape. No dataset in the current results
triggers this branch: all three, including TinyImageNet on its SimCLR
embedding, are bimodal, so this fallback is documented here as a
robustness mechanism of the method rather than illustrated with a live
example from the current results.

\begin{algorithm}[t]
\caption{Adaptive Gate (Class-Balanced Threshold with Receiver-Threshold Fallback)}
\label{alg:coverage_fallback}
\begin{algorithmic}[1]
\Require Peakedness scores $\{c_i\}_{i \in \mathcal{U}}$; predicted class $\hat{y}_i$ for each $i \in \mathcal{U}$; per-class mean belief mass $\{\mu_c\}$; class count $M$; receiver threshold $\tau_r$
\Ensure Distillation-admitted set $\mathcal{T}_\mathcal{U} \subseteq \mathcal{U}$
\State $(\tau, b) \leftarrow \mathrm{ModeThreshold}(\{c_i\})$ \Comment{Algorithm~\ref{alg:mode_threshold}}
\If{$\neg b$} \Comment{no separated high-confidence mode: unimodal case}
    \State $\tau \leftarrow \tau_r$
\EndIf
\State $\bar{\mu}_c \leftarrow \mu_c / \max_{c'} \mu_{c'}$ for each class $c$ \Comment{Eq.~\ref{eq:class_threshold}, main paper}
\State $\tau_c \leftarrow \max(\bar{\mu}_c\,\tau,\; 1/M)$ for each class $c$ \Comment{Eq.~\ref{eq:class_threshold}, main paper}
\State $\mathcal{T}_\mathcal{U} \leftarrow \{i \in \mathcal{U} \mid c_i > \tau_{\hat{y}_i}\}$
\State \Return $\mathcal{T}_\mathcal{U}$
\end{algorithmic}
\end{algorithm}

The fallback introduces no per-dataset switches: the same rule is applied
uniformly, and in the current results every dataset's peakedness
distribution is bimodal, so Algorithm~\ref{alg:coverage_fallback} always
takes the mode-based branch ($b=\mathrm{true}$), a consequence of the
data, not of a manual setting.


\subsubsection{Class-Balanced Threshold Rescaling}
\label{app:class_balanced_threshold}

Algorithm~\ref{alg:mode_threshold} returns a single scalar threshold (its
bimodal indicator is used only to select the base threshold, per the
Unimodal Fallback discussion above), which
Section~\ref{subsec:training} of the main paper rescales per class before applying it. The class
statistic $\mu_c$ is the mean normalized belief mass of column $c$, but
averaged only over that column's top $n = \lfloor N/M \rfloor$ entries, the
class size under the balanced-class assumption, rather than over all $N$
points, since points the matrix assigns to other classes would dilute the
estimate with mass that has nothing to do with class $c$:
\begin{equation}
\begin{split}
    \bar{\mu}_c &= \mu_c \big/ \max_{c'} \mu_{c'},
    \quad
    \tau_c = \max\!\left(\bar{\mu}_c\,\tau,\; 1/M\right) \\
    \mu_c &= \tfrac{1}{n}\textstyle\sum_{i \in \mathrm{Top}_n(c)} \tilde{V}_{i,c}
\end{split}
\tag{S1}
\label{eq:class_threshold_supp}
\end{equation}
where $\mathrm{Top}_n(c)$ is the set of the $n$ points with the largest
$\tilde{V}_{i,c}$. A point $x_i$ is admitted when $c_i > \tau_{\hat{y}_i}$, with
$\hat{y}_i = \arg\max_c \tilde{V}_{i,c}$. Since $\bar{\mu}_c \le 1$ by
construction, $\bar{\mu}_c\,\tau \le \tau$. Whenever $\tau \geq 1/M$ ---
true of every base threshold observed in the current results --- $\tau$
therefore bounds every class threshold from above without an explicit upper
clamp: the rescaling only ever \emph{loosens} the gate relative to the base,
and only for classes below the mass leader, which is gated at $\tau$ exactly.

The floor deserves comment, because it is not a free parameter. A peakedness of
$1/M$ is the score of a uniform belief vector, that is, of a point carrying no
class information at all. Flooring $\tau_c$ there states that no point whose
belief is at or below chance may be distilled from, however little mass its
class holds. This is the same bar the method already uses twice: $1/M$ is the
symmetric Dirichlet prior $\alpha_0$ of Section~\ref{subsec:veracity_matrix}
of the main paper,
and it is the sender condition $c_i > 1/M$ of Veracity Propagation
(Section~\ref{subsec:propagation} of the main paper). The floor binds only for classes with
$\bar{\mu}_c < 1/(M\tau)$, so in practice it is a degeneracy guard against a
near-empty class rather than an active part of the mechanism.

Two further properties are worth making explicit. First, $\bar{\mu}_c$ is a
property of the belief geometry alone, computed before any classifier exists,
so unlike \FlexMatch's~\citep{zhang2021flexmatch} learning-status estimate it
requires neither a trained model nor a warm-up period. This is what makes the
curriculum usable at the first distillation step, in a regime where a
classifier is not yet calibrated enough to rank classes by difficulty at all.
Second, because the statistic is a marginal over the whole unlabeled pool
rather than a per-point quantity, it costs one pass over
$\tilde{\mathbf{V}}$ and is recomputed only when the matrix changes.


\section{Experimental Protocol and Reproducibility}
\label{app:protocol}


\subsection{Implementation Details and Reproducibility}
\label{app:implementation}

\paragraph{Self-supervised pretraining.}
All frozen embeddings (BYOL, SimCLR, Barlow Twins) share a ResNet-18~\citep{he2016deep} backbone and are trained with the \texttt{solo-learn} library~\citep{turrisi2022solo}, using each method's official default configuration on the target dataset's own unlabeled split, unmodified. No SSL-side hyperparameter is tuned as part of this work: \VAST\ and every Track~1 baseline consume the resulting frozen 512-d backbone output (the projector and predictor used to define each pretraining objective are discarded after pretraining, following standard practice for all three methods), so none of the reported differences between methods can be attributed to per-method SSL tuning.

\paragraph{Implementation details.}
The classifier $f_\theta$ is a two-layer MLP: input (512-d, the frozen backbone output) $\to$ linear(256) $\to$ ReLU $\to$ linear(num\_classes), with num\_classes $=100$ for CIFAR-100 and ImageNet100 and $200$ for TinyImageNet.
We train with Adam ($\beta_1{=}0.9$, $\beta_2{=}0.999$, $\epsilon{=}10^{-8}$, weight decay $3{\times}10^{-2}$) at a fixed learning rate of $10^{-3}$ for 300 epochs.
Batch size is 200 for CIFAR-100 and ImageNet100, and 100 for TinyImageNet; batches are constructed with balanced distillation sampling, guaranteeing at least $\max(1, \lfloor\text{batch\_size}/8\rfloor)$ labeled points per batch.
The distillation loss weight is selected empirically per dataset: $\lambda{=}64$ for CIFAR-100 and ImageNet100, and $\lambda{=}32$ for TinyImageNet.
The distillation gate is applied per run: in the current results, it selects the class-balanced mode-based threshold on all three datasets, including TinyImageNet, whose peakedness distribution is bimodal under its SimCLR embedding (the ``Unimodal Fallback'' part of \app~\ref{app:mode_threshold} describes the fallback that would engage instead on a low-purity, unimodal embedding, not triggered here). Per-class thresholds are floored at chance level, $\tau_c = \max(\bar{\mu}_c\tau, 1/M)$, on all datasets, as derived under ``Class-Balanced Threshold Rescaling'' in the same \app. Across all datasets, the Dirichlet prior is set to $\alpha_0{=}1/M$ ($0.01$ for CIFAR-100 and ImageNet100, $0.005$ for TinyImageNet). The connectivity criterion selects $k{=}13$ for CIFAR-100 (BYOL), $k{=}5$ for ImageNet100 (BYOL), and $k{=}3$ for TinyImageNet (SimCLR).

\paragraph{Computing infrastructure.}
All experiments were run on a shared university cluster under Linux.
\VAST, the graph-based baselines (\Poisson, \Laplace, \CS, \Iscen, \LaplacianShot), and the BYOL/SimCLR pretraining runs used NVIDIA A10 (24\,GB) or NVIDIA RTX~6000 (24\,GB) GPUs;
the \mbox{*Match} baselines (\FlexMatch, \FreeMatch, \CoMatch, \SimMatch) used NVIDIA A10 GPUs.
The software stack was Python~3.9, PyTorch~2.x, NumPy, SciPy, and scikit-learn; $k$-NN graph construction used FAISS.
Wall-clock times reported in Section~\ref{subsec:main_results} of the main paper and \app~\ref{app:timing} were measured on A10 hardware.

\paragraph{Reproducibility.}
All results are reported as mean $\pm$ SEM (standard error of the mean) over 5 independent runs with seeds 0--4 on CIFAR-100, 6 runs with seeds 0--5 on ImageNet100, and 5 runs with seeds 0--4 on TinyImageNet for \VAST, \Poisson, \Laplace, \CS, \LaplacianShot, and \Iscen (the remaining TinyImageNet baselines MLP labeled-only, \FlexMatch, \FreeMatch are at 3 runs or fewer, seeds 0--2). One exception: on TinyImageNet, the reported dispersions for \VAST\ and these same frozen-embedding baselines (Table~1 of the main paper, and every TinyImageNet figure and table in this supplement that reproduces or sweeps around those runs) are sample standard deviations over the 5 labeled-set draws, not standard errors; CIFAR-100 and ImageNet100 dispersions throughout are SEM as stated above.
Labeled sets are shared across all methods per seed, enabling paired comparison between \VAST and each baseline on identical data; statistical significance is assessed with two-sided paired $t$-tests on shared seeds.


\subsection{Cross-Dataset Table: Full Protocol Details}
\label{app:crossdataset_protocol}

Graph configurations: $k{=}50$ for \Poisson, \CS, \LaplacianShot, and \Iscen on all three datasets; \Laplace uses $k{=}20$ on CIFAR-100 and $k{=}50$ on TinyImageNet and ImageNet100 (best-performing configuration per baseline).
Seed convention: unless noted otherwise, entries follow the per-dataset protocol of the main paper (CIFAR-100: 5 seeds; ImageNet100: 6; TinyImageNet: 5), and entries reported without a SEM are single runs.
Exception --- \FreeMatch: 2 seeds (0--1) on CIFAR-100, and on TinyImageNet and ImageNet100 2 seeds (0--1) at 1--2 labels per class and 1 seed (0) at 4 labels per class; the 2-seed means are provisional estimates.
Training duration was fixed in advance as a wall-clock compute budget per run, not extended or cut short based on interim results (CIFAR-100 completed the full $2^{20}{\approx}1048$K schedule within its budget; TinyImageNet runs reached 570--1048K and ImageNet100 431--835K iterations within theirs). This is a statement about \emph{when training stops}, not about which checkpoint's accuracy is reported: within each fixed-budget run, checkpoints are saved periodically and the best one is reported, per the checkpoint-selection convention detailed in the Track~2 protocol below; the two practices are independent and both apply throughout.
\VAST results use the final graph configuration described in the main paper; TinyImageNet \VAST dispersions (sample SD, see above) are computed over 5 independent labeled-set draws on the SimCLR embedding (per-seed accuracies at 1/cls: 16.54/18.12/18.9/16.21/18.76).


\subsubsection{Track 2 Table: Full Protocol Details}
\label{app:track2_protocol}

All \mbox{*Match} baselines use the official USB benchmark
codebase~\citep{usb2022} with the WRN-28-8 from-scratch
protocol (no pre-trained initialization).
At 1 and 2 labels per class, we ran the USB codebase locally
and report best-checkpoint accuracy, following USB's own
evaluation convention; at 4 labels per class, \mbox{*Match}
results are taken directly from the official USB
repository.\footnote{\url{https://github.com/microsoft/Semi-supervised-learning}}
All three budgets therefore share the same backbone,
codebase, and checkpoint-selection discipline (best checkpoint).

\CoMatch and \SimMatch are single run (seed~0) at 1 and 2
labels per class; their 4-labels-per-class entries are taken
directly from the USB repository.
\FreeMatch uses 2 seeds (0--1) on CIFAR-100 (completing
the full $2^{20}{\approx}1048$K schedule), 2 seeds (0--1) at
1--2 labels per class on TinyImageNet and ImageNet100,
and 1 seed (0) at 4 labels per class on those datasets;
the 2-seed means are provisional estimates.
Training iterations: \FlexMatch (WRN-28-2) 675K;
\FlexMatch (WRN-28-8) 934K/944K;
\FreeMatch (WRN-28-8) $2^{20}{\approx}1048$K;
\CoMatch 484K (100 labels), 426K (200 labels);
\SimMatch 590K (100 labels), 606K (200 labels).
Results across all label budgets (1/2/4 per class)
also appear in the cross-dataset table.

\paragraph{\ExMatch.} Unlike the other Track~2 baselines, \ExMatch~\citep{kim2024exmatch} has no public implementation, so it is not reproduced locally: the $32.49_{\pm1.3}$/$43.71_{\pm0.7}$/$51.13_{\pm1.7}$ figures in Table~\ref{tab:track2} are quoted directly from the original paper at matching label budgets (100/200/400 labels, CIFAR-100) and are not independently verified under our labeled-set draws, seeds, or evaluation protocol. It is included for completeness as the one prior method in Table~\ref{tab:track2} explicitly designed for the scarce-label regime, not as a like-for-like reproduction; the main paper's headline margin against the strongest \emph{reproduced} \mbox{*Match} entry is reported both with and without \ExMatch\ admitted (Section~\ref{subsec:main_results}) precisely because of this asymmetry in verification.


\subsection{Backbone and Architecture Parameter Counts}
\label{app:params}

Table~\ref{tab:params} compares parameter counts across every architecture used in Track~1 and Track~2, to make explicit that the compared backbones are not size-matched.

\begin{table}[tb]
\centering
\scriptsize
\begin{tabular}{@{}lc@{}}
\toprule
Architecture & Parameters \\
\midrule
ResNet-18 backbone (\VAST/BYOL) & ${\sim}11.2$M \\
BYOL projector (2-layer MLP, 512$\to$256$\to$256) & ${\sim}0.2$M \\
BYOL predictor (same architecture) & ${\sim}0.2$M \\
\textbf{Full BYOL network (backbone + projector + predictor)} & \textbf{${\sim}11.6$M} \\
Labeled-only MLP classifier head & ${\sim}0.16$--$0.18$M \\
\midrule
WRN-28-2 (\FlexMatch) & ${\sim}1.47$M \\
WRN-28-8 (\FlexMatch, \FreeMatch, \CoMatch, \SimMatch) & ${\sim}23.4$M \\
\bottomrule
\end{tabular}
\caption{Parameter counts by architecture.}
\label{tab:params}
\end{table}

The ResNet-18 parameter count follows the standard architecture of \citet{he2016deep}. Our BYOL implementation's projector and predictor use a 256-d hidden layer (mapping the 512-d ResNet-18 representation to a 256-d hidden layer and then to a 256-d embedding), smaller than the 4096-d hidden layer used in the original BYOL paper's \citep{grill2020bootstrap} ImageNet/ResNet-50 setup.
WRN-28-8 has ${\sim}16\times$ more parameters than WRN-28-2, and is roughly $2\times$ larger than the full BYOL network used by \VAST; WRN-28-2 is smaller than the full BYOL network. Neither \mbox{*Match} backbone is size-matched to \VAST's.


\section{Additional Experimental Results}
\label{app:additional_results}


\subsection{Track 1: Full Significance Testing and \PoissonMBO}
\label{app:track1_significance}

Section~4.2 of the main paper reports that \VAST leads \Poisson Learning at every budget on all three datasets. The full per-budget statistics are as follows.
At 100 labels \VAST (39.2) leads over \Poisson (38.6) on 4/5 seeds, but the margin is not yet significant (paired $t$-test, $p{=}0.36$); at 200 labels \VAST leads 47.8 vs.\ 45.1 significantly ($p{=}0.016$, 5/5 seeds), and at 400 labels 54.4 vs.\ 50.1 ($p{=}0.0023$, 5/5 seeds).
Against the other four Track~1 baselines on CIFAR-100 (5 seeds throughout), \VAST significantly leads \Laplace at every budget ($p{=}0.0025$, $0.0006$, $p{<}0.0001$ at 100/200/400 labels; 5/5 seeds throughout), \CS at every budget ($p{=}0.0002$, $0.0090$, $0.0071$; 5/5 seeds throughout), and \LaplacianShot at every budget ($p{=}0.0001$, $0.0004$, $0.0001$; 5/5 seeds throughout). Against \Iscen, the margin is significant at 400 labels ($p{=}0.0079$, 5/5 seeds) but borderline at 100 and 200 labels under the standard 5-seed protocol ($p{=}0.063$, 4/5 seeds; $p{=}0.056$, 5/5 seeds); for this comparison only, a sixth labeled-set draw was added after the initial 5-seed result, resolving both budgets ($n{=}6$: 39.19 vs.\ 37.04, $p{=}0.0247$, 5/6 seeds at 100 labels; 47.87 vs.\ 45.74, $p{=}0.0244$, 6/6 seeds at 200 labels). This is the only comparison in this appendix using 6 seeds on CIFAR-100; every other CIFAR-100 cell, including the other three baselines above, uses the standard 5.

On ImageNet100, \VAST leads over \Poisson at every budget (34.9 vs.\ 31.7 at 100 labels, 43.1 vs.\ 38.5 at 200, 49.3 vs.\ 42.0 at 400), and the margin is significant at all three budgets (100 labels: $p{=}0.0049$; 200 labels: $p{=}0.0199$; 400 labels: $p{<}0.0001$). The comparison is moreover sign-consistent at the seed level: \VAST wins on every individual seed at every ImageNet100 budget (6/6 seeds), a statement that holds independently of the paired-test analysis.
Against the other four Track~1 baselines on ImageNet100, \VAST significantly leads \CS at every budget ($p{<}0.0001$, $0.0022$, $0.0002$; 6/6 seeds throughout) and \Iscen at every budget ($p{=}0.0018$, $0.0202$, $0.0002$; 6/6 seeds throughout), both at the full 6-seed protocol. \VAST\ likewise significantly leads \Laplace at every budget ($p{=}0.0004$, $0.0004$, $0.0002$; 6/6 seeds throughout) and \LaplacianShot at every budget ($p{=}0.0006$, $0.0017$, $0.0002$; 6/6 seeds throughout).

On TinyImageNet (SimCLR embedding; see main paper Section~4.1), \VAST leads over \Poisson by a significant margin at every budget (17.71 vs.\ 16.60 at 200 labels, $p{=}0.028$, 4/5 seeds; 22.83 vs.\ 21.06 at 400, $p{=}0.016$, 5/5 seeds; 27.55 vs.\ 24.36 at 800, $p{=}0.0065$, 5/5 seeds). Against the other TinyImageNet baselines: \VAST significantly leads \Laplace and \Iscen at all three budgets ($p<0.003$ throughout) and \CS at 200 and 400 labels ($p{=}0.0011$ and $p{=}0.0142$), but the margin over \CS at 800 labels is \emph{not} significant ($+0.66$ pts, $p{=}0.41$, 4/5 seeds), the one baseline comparison on TinyImageNet that is not a confirmed win. \VAST also significantly leads \LaplacianShot at every budget (17.71 vs.\ 14.60 at 200 labels, $p{=}0.0006$, 5/5 seeds; 22.83 vs.\ 20.24 at 400, $p{=}0.001$, 5/5 seeds; 27.55 vs.\ 25.37 at 800, $p{=}0.0005$, 5/5 seeds).

We additionally ran the MBO variant of \Poisson Learning (\PoissonMBO) on CIFAR-100. It underperforms standard \Poisson Learning at 100 and 200 labels ($35.3\%$ vs.\ $38.6\%$ and $44.9\%$ vs.\ $45.1\%$, respectively) and overtakes it by roughly one point at 400 labels ($51.1\%$ vs.\ $50.1\%$); we did not run a significance test on these differences. We omit \PoissonMBO from the main paper's cross-dataset table primarily because it could not be run at all on TinyImageNet or ImageNet100, since the reference implementation does not scale to datasets of this size and runs failed to terminate within a practical compute budget, so including CIFAR-100-only figures would leave the cross-dataset table inconsistent with every other row, which reports all three datasets.

All $p$-values above are reported per cell without multiple-comparison correction; we treat them as descriptive rather than as a family-wise test. The correction-free evidence for \VAST's advantage is the seed-level sign-consistency (\VAST\ wins 5/5 or 6/6 seeds at most budgets), which does not depend on any significance threshold.


\subsection{Propagation Termination}
\label{app:prop_convergence}

Termination (Section~3.4 of the main paper) is fast and consistent: roughly 12--13 rounds on CIFAR-100 (BYOL), 22--23 on ImageNet100 (BYOL), and 24--27 on TinyImageNet (SimCLR). The connectivity criterion selects $k=5$ on ImageNet100 and $k=3$ on TinyImageNet vs.\ $k=13$ on CIFAR-100, so evidence traverses longer paths before stabilizing on the two smaller-$k$ datasets.


\subsection{Receiver Threshold Sensitivity}
\label{app:receiver_ablation}

The receiver threshold $\tau_r$ is the only fixed hyperparameter in the propagation stage (Section~3.4 of the main paper).
Sweeping $\tau_r \in \{0.6, 0.7, 0.75, 0.8, 0.9\}$ on CIFAR-100 (5 seeds per cell) leaves \VAST stable at every budget: the sweep range sits within one seed-level SEM at 100 labels, and is discernible above SEM but small (under 0.8 pts) at 200 and 400 labels, with no off-default setting a clear outlier and $\tau_r = 0.75$ (our fixed default) tracking the best or near-best mean throughout (Figure~\ref{fig:tau_r_sensitivity} in Section~\ref{subsec:propagation} of the main paper). The mild decline at $\tau_r = 0.9$ reflects nearly-converged points continuing to receive second-hand evidence, delaying propagation's termination.
The same sweep on TinyImageNet (SimCLR, 5 seeds per cell) at 200/400/800 labels shows the same pattern: ranges stay small (under 1.3 pts), no setting is a clear outlier, and $\tau_r = 0.75$ tracks the best mean at every budget (Figure~\ref{fig:tau_r_sensitivity_tinyimagenet}). Its 800-label default cell reproduces the canonical headline result ($27.55_{\pm1.61}$, Table~\ref{tab:crossdataset} of the main paper) rather than an independent measurement at that setting, so that column is not a fully like-for-like comparison across all five settings.

\begin{figure}[t]
\centering
\includegraphics[width=0.9\linewidth]{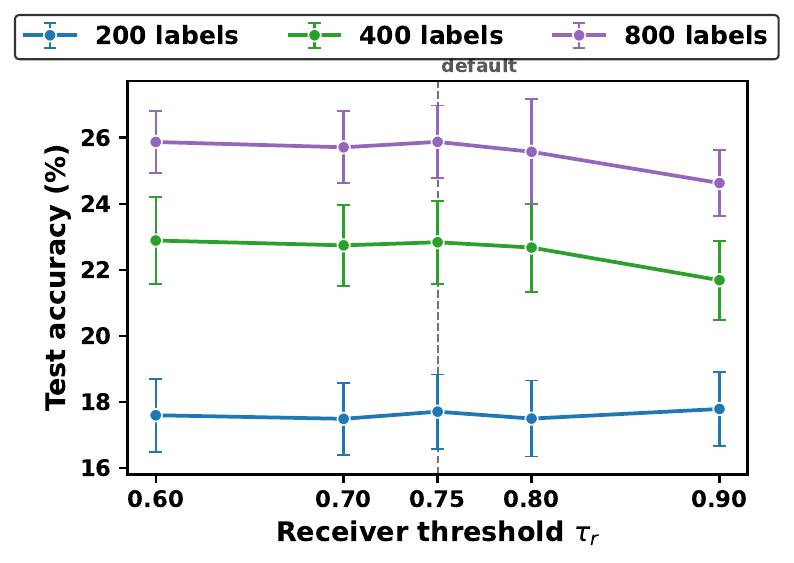}
\caption{Receiver-threshold sensitivity on TinyImageNet (SimCLR). Test accuracy
(mean $\pm$ SD over 5 seeds) as a function of $\tau_r$ at 200, 400, and 800 labels.
The dashed line marks the default $\tau_r = 0.75$.}
\label{fig:tau_r_sensitivity_tinyimagenet}
\end{figure}


\subsection{Dirichlet Prior ($\alpha_0$) Sensitivity}
\label{app:alpha0_ablation}

The Dirichlet smoothing prior $\alpha_0$ (Section~3.2 of the main paper) is a fixed hyperparameter entering every entry of the Veracity Matrix. It is set a priori to $\alpha_0 = 1/M$ ($0.01$ on CIFAR-100), never tuned. As in the receiver-threshold and distillation-weight sweeps, the default-$\alpha_0$ cell reproduces the canonical headline result (Table~\ref{tab:crossdataset} of the main paper) rather than an independent measurement at that setting. To check that this a priori value does not sit on a sharp optimum, we swept $\alpha_0 \in \{0.0075, 0.0095, 0.01, 0.015, 0.025\}$ on CIFAR-100 (5 seeds per cell), a $3.3\times$ band centred on the default. Mean accuracy spans $38.83$--$39.36$\% at 100 labels (range $0.53$ pts), $47.44$--$47.80$\% at 200 labels (range $0.36$ pts), and $53.89$--$54.41$\% at 400 labels (range $0.52$ pts). As with the receiver threshold $\tau_r$ (\app~\ref{app:receiver_ablation}), the range at 100 labels sits within a single seed-level SEM ($0.71$--$0.83$), while at 200 and 400 labels it is discernible above seed-to-seed noise but small in absolute terms. The default is within noise of the best setting at every budget rather than optimal at any of them (Figure~\ref{fig:alpha0_sensitivity}).

\begin{figure}[t]
\centering
\includegraphics[width=0.9\linewidth]{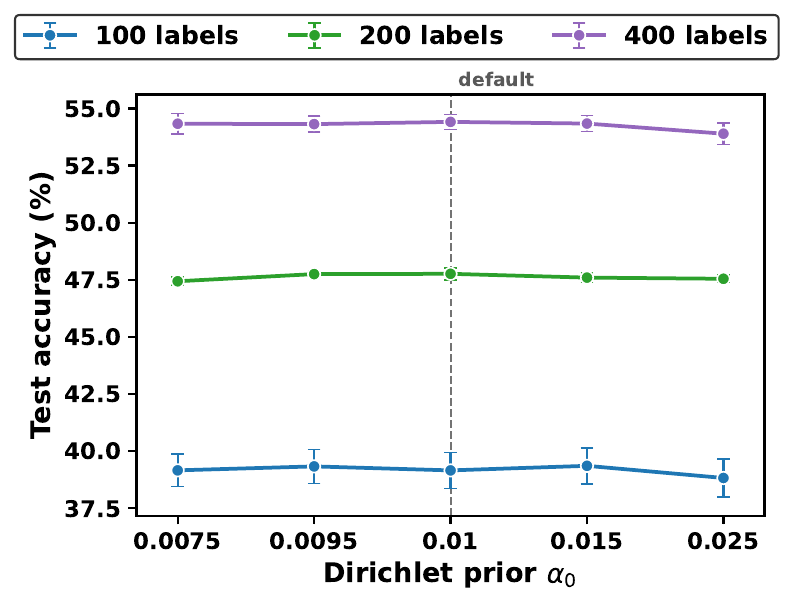}
\caption{Dirichlet prior ($\alpha_0$) sensitivity on CIFAR-100 (BYOL), in a narrow band around the a priori value $\alpha_0 = 1/M = 10^{-2}$ (dashed line). Mean $\pm$ SEM over 5 seeds.}
\label{fig:alpha0_sensitivity}
\end{figure}

We separately checked $\alpha_0$'s local sensitivity on TinyImageNet (SimCLR, 5 seeds per cell) at label budgets of 200/400/800, sweeping $\alpha_0 \in \{0.001, 0.0025, 0.005, 0.0075, 0.01\}$, a band centred on the dataset's a priori default $\alpha_0 = 1/M = 0.005$ (Section~3.2 of the main paper; note this differs from CIFAR-100's $1/M=0.01$, since TinyImageNet has 200 classes). Mean accuracy spans $16.89$--$17.71$\% at 200 labels (range $0.82$ pts), $22.28$--$22.90$\% at 400 labels (range $0.62$ pts), and $27.40$--$28.06$\% at 800 labels (range $0.66$ pts), comparable in magnitude to the CIFAR-100 ranges (\app~\ref{app:alpha0_ablation}, CIFAR-100 paragraph above). As with CIFAR-100, the default is within noise of the best setting at every budget rather than optimal at any of them (Figure~\ref{fig:alpha0_sensitivity_tinyimagenet}).

\begin{figure}[t]
\centering
\includegraphics[width=0.9\linewidth]{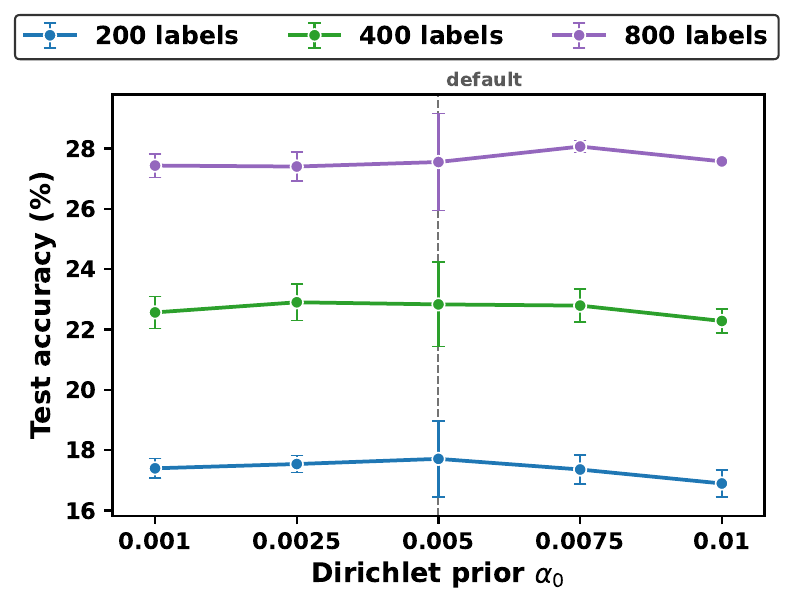}
\caption{Dirichlet prior ($\alpha_0$) sensitivity on TinyImageNet (SimCLR), in a narrow band around the a priori value $\alpha_0 = 1/M = 5\times10^{-3}$ (dashed line). Mean $\pm$ SD over 5 seeds.}
\label{fig:alpha0_sensitivity_tinyimagenet}
\end{figure}


\subsection{Distillation Weight ($\lambda$) Sensitivity}
\label{app:lambda_ablation}

The distillation weight $\lambda$ (Section~3.5 of the main paper) is a fixed hyperparameter scaling the distillation term relative to cross-entropy. (The $\lambda \in [1, 256]$ range quoted in Section~3.5 of the main paper refers to the $16$--$256$ band swept below; the $\lambda=1$ probe reported at the end of this section sits outside that band and is treated separately.) Sweeping $\lambda \in \{16, 32, 64, 128, 256\}$ on CIFAR-100 (5 seeds per cell) gives mean accuracies of 38.97--39.2\% at 100 labels (range 0.23 pts), 47.44--47.80\% at 200 labels (range 0.36 pts), and 53.91--54.4\% at 400 labels (range 0.49 pts); across the sweep, seed-level standard errors range 0.75--0.90\% at 100 labels, 0.23--0.30\% at 200 labels, and 0.31--0.40\% at 400 labels. The $\lambda=64$ cell reproduces the canonical headline result (Table~\ref{tab:crossdataset}) rather than an independent measurement at this setting, consistent with the same convention used for the other sensitivity sweeps (\app~\ref{app:receiver_ablation}, \app~\ref{app:alpha0_ablation}). As with the receiver threshold $\tau_r$, the sweep range at 100 labels remains within the seed-level SEM, while at 200 and 400 labels the range is discernible above seed-to-seed noise but small in absolute terms. Within this $16\times$ range, \VAST is stable to within under a point of accuracy, and $\lambda=64$ (our fixed default) tracks the best mean at every budget. We additionally checked $\lambda=1$, well below the swept range, specifically to probe where this robustness ends rather than to extend the stability claim further: it shows a real drop rather than continued stability, with accuracy of 37.93/44.91/50.15\% at 100/200/400 labels, 1.3--4.3 points below the default depending on budget. \VAST\ is therefore robust across the wide $16$--$256$ band, a $16\times$ range, but this robustness is not unconditional, since sufficiently small $\lambda$ under-weights the distillation term enough to measurably degrade accuracy (Figure~\ref{fig:lambda_sensitivity}).

\begin{figure}[t]
\centering
\includegraphics[width=0.9\linewidth]{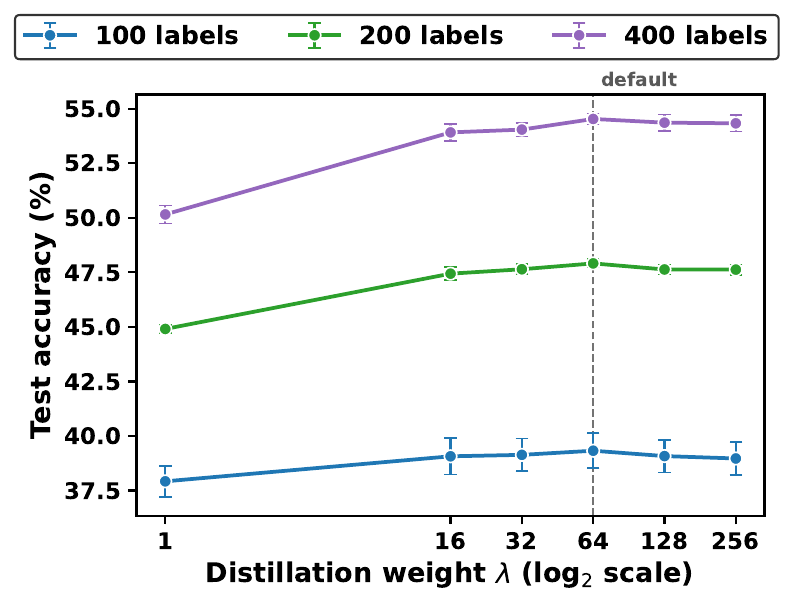}
\caption{Distillation weight ($\lambda$) sensitivity on CIFAR-100 (BYOL), log$_2$ scale; dashed line marks the default $\lambda = 64$. Mean $\pm$ SEM over 5 seeds.}
\label{fig:lambda_sensitivity}
\end{figure}


\subsection{Cross-Dataset Embedding Geometry}
\label{app:embedding_geometry}

This section covers four topics referenced from the main paper:
\textbf{Intrinsic dimensionality and connectivity $k$} (referenced from Section~3.3, ``Connectivity-based $k$-NN graph'');
\textbf{Degree distribution} (referenced from Section~3.3, ``Density normalization'');
\textbf{$k$-NN edge purity} (referenced from Section~3.3, ``Density normalization''); and the
\textbf{TinyImageNet embedding choice: BYOL vs.\ SimCLR} (referenced from Section~4.1, ``Embeddings'').

\paragraph{Intrinsic dimensionality and connectivity $k$.}
The connectivity-selected $k$ (Section~3.3 of the main paper) tracks structural properties of the feature space it is computed on. CIFAR-100 and ImageNet100 use BYOL embeddings throughout; TinyImageNet uses SimCLR instead, for the embedding-quality reasons given in the main paper (Section~4.1). Table~\ref{tab:pca_dim} reports the PCA participation ratio (PR), a standard scalar measure of effective dimensionality, computed on each dataset's feature covariance. Within the BYOL pair, CIFAR-100's embeddings occupy more than twice the effective dimensions of ImageNet100, matching the larger connectivity $k$ it requires ($k=13$ vs.\ $k=5$). We do not extend this dimensionality comparison across the BYOL/SimCLR boundary to TinyImageNet: PR is a joint property of a dataset \emph{and} its embedding, so a cross-encoder PR comparison would conflate the two factors; consistent with this, TinyImageNet's SimCLR PR (75.5) falls \emph{between} ImageNet100's and CIFAR-100's BYOL values despite TinyImageNet keeping the smallest connectivity $k$ ($k=3$) of the three datasets; the monotonic PR-to-$k$ relationship that holds within the BYOL pair does not hold once the embedding itself changes. TinyImageNet's $k=3$ is justified on its own terms, by the same label-free connectivity criterion applied to its own embedding, independent of the BYOL-only dimensionality argument.

\begin{table}[tb]
\centering
\begin{tabular}{lccc}
\toprule
 & CIFAR-100 & ImageNet100 & TinyImageNet \\
\midrule
Encoder & BYOL & BYOL & SimCLR \\
$N$ & 50{,}000 & 130{,}000 & 100{,}000 \\
PR & 138.1 & 60.6 & 75.5 \\
$n$@90\% & 252 & 164 & 214 \\
$n$@95\% & 339 & 256 & 306 \\
$n$@99\% & 460 & 425 & 441 \\
\bottomrule
\end{tabular}
\caption{PCA intrinsic dimensionality. CIFAR-100/ImageNet100 on BYOL embeddings; TinyImageNet on SimCLR. $N$ is the labeled+unlabeled pool size; PR is the participation ratio; the last three rows report the number of principal components needed to explain the given fraction of variance.}
\label{tab:pca_dim}
\end{table}

\paragraph{Degree distribution.}
The sparse $k$-NN graph's degree distribution is also relevant to the density-normalization step (Section~3.3 of the main paper): a right-skewed, hub-prone degree distribution is precisely the condition under which a polynomial density correction (as in the original Coifman--Lafon anisotropic normalization) risks suppressing hub-adjacent edges enough to fracture manifold connectivity, motivating the paper's logarithmic variant instead. Table~\ref{tab:degree_stats} reports hard degree (neighbor count) and soft degree (sum of RBF neighbor weights) statistics at each dataset's connectivity-selected $k$, computed over the full labeled+unlabeled pool, using the same embeddings as the main results (BYOL for CIFAR-100/ImageNet100, SimCLR for TinyImageNet). All three show the same qualitative pattern: a mean degree well below the tail, with the 99th-percentile point connecting to roughly $2.9$--$3.4\times$ the mean number of neighbors, and isolated hub points reaching $9.8$--$25.6\times$ the mean. The effect is most pronounced on CIFAR-100, whose higher effective dimensionality (Table~\ref{tab:pca_dim}) and larger connectivity-selected $k$ produce both a higher mean degree and a longer tail. TinyImageNet's degree statistics under BYOL also showed the same right-skewed shape, indicating the phenomenon holds under both encoders and is a property of the sparse graph construction rather than specific to either one. Figure~\ref{fig:degree_hist} visualizes the full distributions.

\begin{table*}[tb]
\centering
\small
\setlength{\tabcolsep}{8pt}
\begin{tabular}{llccccccc}
\toprule
& & & \multicolumn{3}{c}{Hard degree} & \multicolumn{3}{c}{Soft degree} \\
\cmidrule(lr){4-6} \cmidrule(lr){7-9}
Dataset & Encoder & $k^*$ & mean & max & p99 & mean & max & p99 \\
\midrule
CIFAR-100 & BYOL & 13 & 20.3 & 518 & 65 & 12.2 & 280.5 & 37.1 \\
ImageNet100 & BYOL & 5 & 8.0 & 78 & 23 & 4.8 & 44.3 & 13.7 \\
TinyImageNet & SimCLR & 3 & 5.1 & 50 & 17 & 3.0 & 28.7 & 9.9 \\
\bottomrule
\end{tabular}
\caption{Sparse-graph degree statistics at each dataset's connectivity-selected $k$: hard degree (neighbor count) and soft degree (sum of RBF neighbor weights), over the full labeled+unlabeled pool. CIFAR-100/ImageNet100 on BYOL embeddings; TinyImageNet on SimCLR. (Cited as \emph{Table 7} in the main paper's Section 3.3.)}
\label{tab:degree_stats}
\end{table*}

\begin{figure*}[t]
\centering
\includegraphics[width=\linewidth]{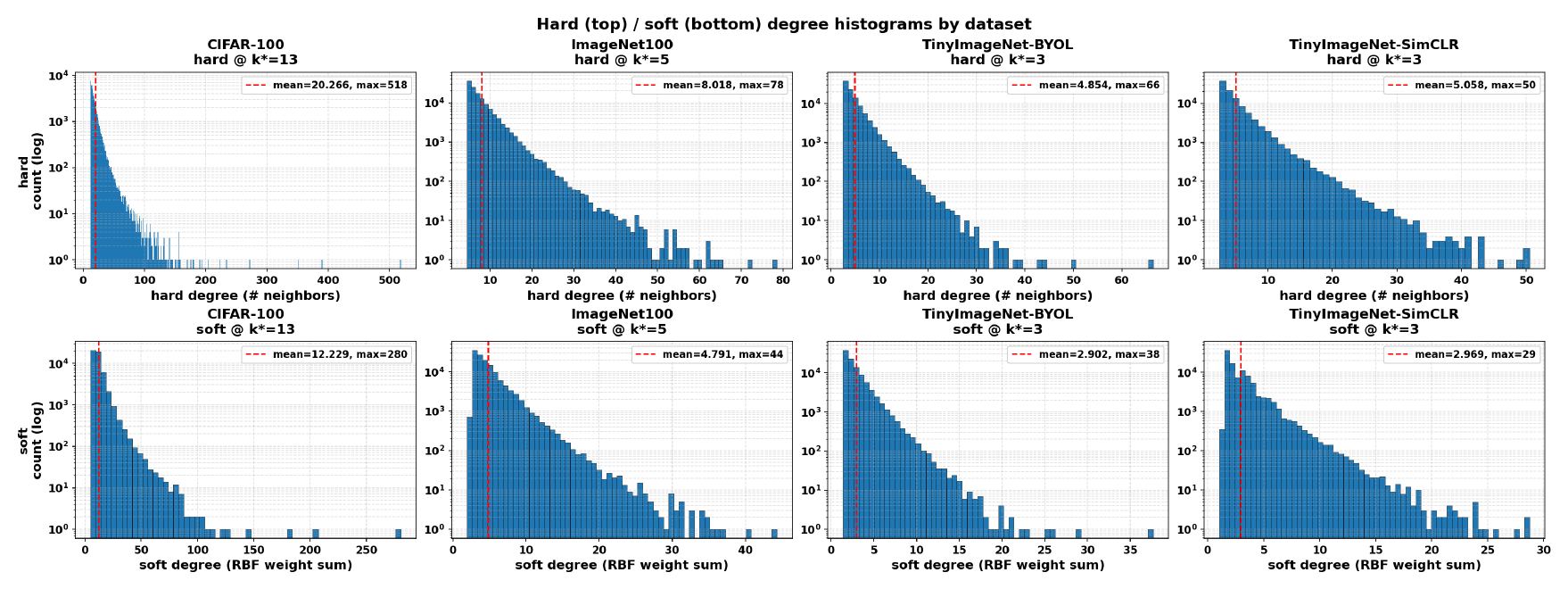}
\caption{Hard degree (top row) and soft degree (bottom row) distributions on the sparse $k$-NN graph at each dataset's connectivity-selected $k$. Dashed lines mark the mean. The additional TinyImageNet--BYOL panel uses the same $k{=}3$ selected on the SimCLR embedding (Table~\ref{tab:degree_stats}), so all panels share a common $k$ per dataset.}
\label{fig:degree_hist}
\end{figure*}

\paragraph{Boundedness and direction of the degree correction.}
Since the denominator $\sqrt{\log(1+d_id_j)}$ of Eq.~4 of the main paper vanishes as $d_i, d_j \to 0$, the correction might appear unbounded; it is not. Because the $k$-NN graph is symmetrized, an edge $(i,j)$ contributes its own weight $K_{ij} \equiv K(x_i,x_j)$ to both endpoint degrees, and all kernel values are positive, so $d_i \geq K_{ij}$ and $d_j \geq K_{ij}$, giving $d_i d_j \geq K_{ij}^2$. As $t \mapsto \sqrt{\log(1+t)}$ is increasing,
\begin{equation*}
\hat{K}_{ij} = \frac{K_{ij}}{\sqrt{\log(1+d_i d_j)}} \;\leq\; \frac{K_{ij}}{\sqrt{\log(1+K_{ij}^2)}} \;=\; g(K_{ij}),
\end{equation*}
where $g(u) = u/\sqrt{\log(1+u^2)}$ is increasing on $(0,1]$ with $g(u) \to 1$ as $u \to 0^+$ and $g(1) = 1/\sqrt{\log 2} \approx 1.20$ (logarithms natural throughout). For the RBF kernel with $2\sigma^2 = 1$ we have $K_{ij} \in (0,1]$, so $\hat{K}_{ij} \leq 1/\sqrt{\log 2}$ on every edge, with no assumption on the degree distribution. The normalization therefore cannot diverge, and it creates or removes no edges, so the connectivity guaranteed by the criterion of Section~3.3 of the main paper is preserved exactly.

The correction is also two-sided rather than purely damping: $\hat{K}_{ij} > K_{ij}$ exactly when $\log(1+d_i d_j) < 1$, i.e.\ when $d_i d_j < e-1 \approx 1.72$. Above that threshold edges are damped, below it they are lifted, subject to the bound above. Both directions serve the motivation of Section~3.3 of the main paper: hub-adjacent edges, whose degree products lie far above the threshold, are suppressed, while sparse-region edges are lifted toward unit weight. At the soft-degree means of Table~\ref{tab:degree_stats} (12.2, 4.8 and 3.0 for CIFAR-100, ImageNet100 and TinyImageNet), typical edges sit well inside the damping regime; the lifting applies to the lower tail of the degree distribution.

The logarithmic form is not specific to this setting. \citet{corso2020principal} introduce degree scalers of the form $(\log(d+1)/\delta)^\alpha$ for graph neural networks, selecting the logarithm over linear scaling on the grounds that the latter amplifies the aggregated signal too aggressively. Our use differs in three respects: the correction is fixed rather than learned, it is applied symmetrically to an edge's two endpoint degrees rather than to a node's incoming aggregate, and it is not normalized by a dataset-level average. The shared element is the choice of $\log(1+d)$ as the moderate alternative to a polynomial or linear penalty.

\paragraph{$k$-NN edge purity.}
Table~\ref{tab:knn_purity} reports $k$-NN edge purity, the fraction of neighbor edges linking same-class points, at matched $k$ across all three datasets. TinyImageNet remains the least separable of the three datasets: at matched $k$ its SimCLR purity is roughly 50--56\% of CIFAR-100/ImageNet100's BYOL purity. This combination of cheap connectivity (small connectivity-selected $k^\star$, Table~\ref{tab:degree_stats}) and comparatively low purity caps the absolute accuracy attainable by \VAST\ and the graph-based baselines alike on TinyImageNet (\Poisson\ at 16.6\% vs.\ \VAST\ at 17.7\% at 1 label/class, against 31.7\%/34.9\% on ImageNet100, Section~4.2 of the main paper): low purity depresses the whole ceiling rather than opening a wider gap between methods, which is why \VAST's margin over \Poisson\ is in fact largest on ImageNet100 rather than TinyImageNet (Section~4.2), not the reverse.

\begin{table}[tb]
\centering
\small
\setlength{\tabcolsep}{4pt}
\begin{tabular}{lccc}
\toprule
$k$ & CIFAR-100 & ImageNet100 & TinyImageNet \\
\midrule
1  & 0.616 & 0.611 & 0.343 \\
3  & 0.581 & 0.572 & 0.312 \\
5  & 0.563 & 0.551 & 0.295 \\
10 & 0.536 & 0.522 & 0.272 \\
15 & 0.520 & 0.505 & 0.258 \\
\bottomrule
\end{tabular}
\caption{$k$-NN edge purity (fraction of same-class neighbor edges) at matched $k$. CIFAR-100/ImageNet100 on BYOL embeddings; TinyImageNet on SimCLR.}
\label{tab:knn_purity}
\end{table}


Figure~\ref{fig:knn_purity} visualizes the full sweep. TinyImageNet's curve sits below the other two at every $k$, but markedly closer than under its earlier BYOL embedding (roughly 50--56\% of the other two datasets, against roughly 33--42\% under BYOL): still the least separable of the three, yet substantially cleaner than BYOL on the same data.

\begin{figure}[tb]
\centering
\includegraphics[width=0.9\linewidth]{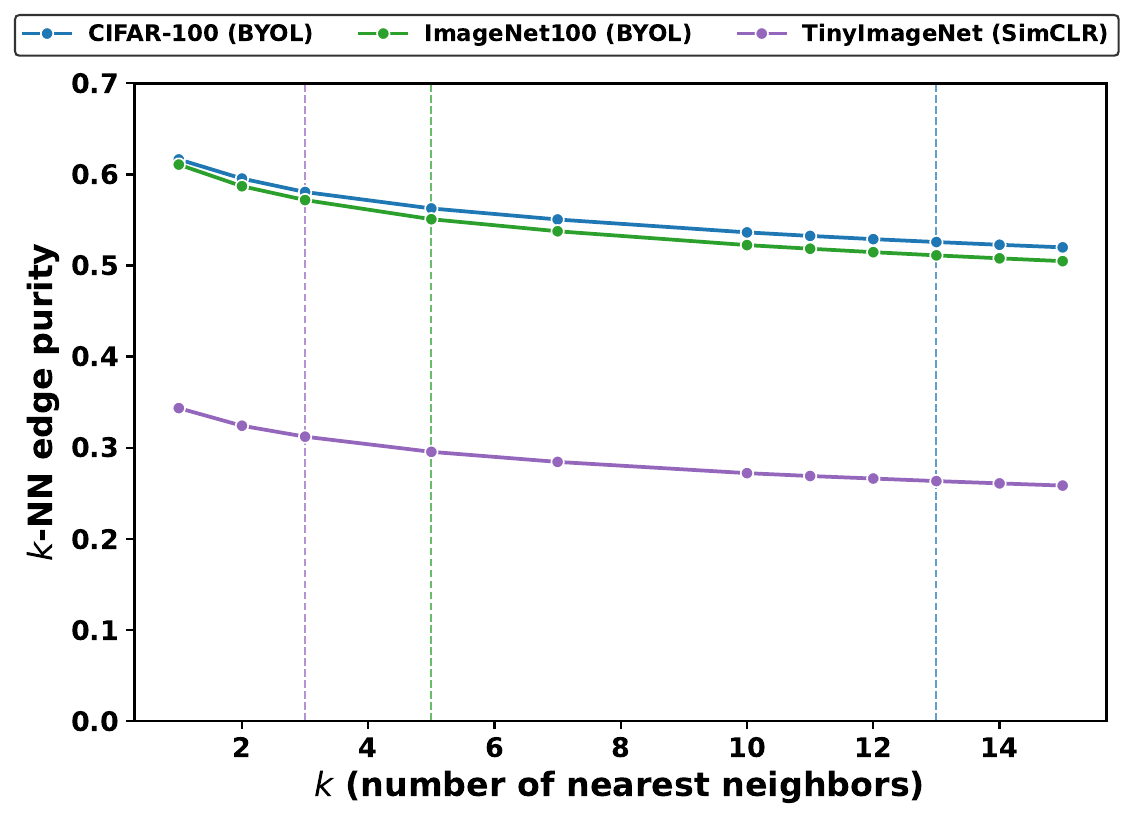}
\caption{$k$-NN edge purity as a function of neighborhood size $k$. Dashed vertical lines mark each dataset's connectivity-selected $k$.}
\label{fig:knn_purity}
\end{figure}


\paragraph{TinyImageNet embedding choice: BYOL vs.\ SimCLR.} Table~\ref{tab:knn_purity} and Figure~\ref{fig:knn_purity} report $k$-NN edge purity across all three datasets, but TinyImageNet appears there only on its SimCLR embedding, since the paper's reported TinyImageNet results all use that embedding (Section~4.1 of the main paper). To make the actual BYOL-vs-SimCLR comparison that motivates the encoder switch explicit, Figure~\ref{fig:tinyimagenet_purity} plots TinyImageNet's own $k$-NN edge purity under both embeddings, on the same $k$ grid. SimCLR purity exceeds BYOL purity by a wide, near-constant margin at every $k$ (0.343 vs.\ 0.256 at $k{=}1$, narrowing only slightly to 0.258 vs.\ 0.172 at $k{=}15$), a purity gap of comparable relative size to the one already reported between TinyImageNet and the other two datasets in Table~\ref{tab:knn_purity}, but here isolating the encoder as the sole variable on a single, fixed dataset.

\begin{figure}[tb]
\centering
\includegraphics[width=0.9\linewidth]{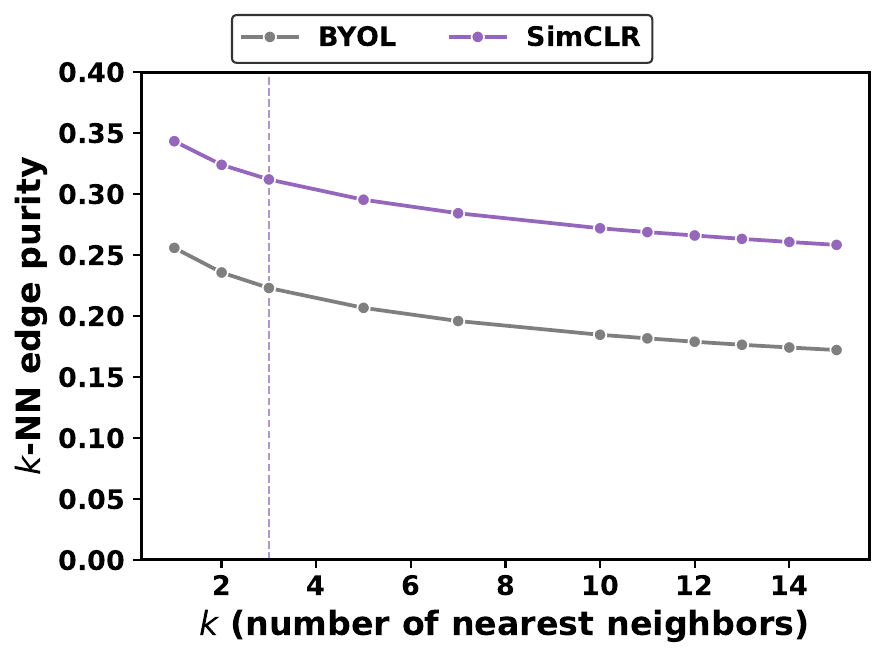}
\caption{TinyImageNet $k$-NN edge purity under BYOL vs.\ SimCLR, as a function of neighborhood size $k$. Dashed line marks the connectivity-selected $k=3$.}
\label{fig:tinyimagenet_purity}
\end{figure}


This purity measure is an oracle diagnostic: it uses every point's true class label, not the handful available to \VAST itself, and is independent of \VAST's own training pipeline; it is corroborated by a classification-based one: we additionally compare 1-NN test accuracy on TinyImageNet under BYOL vs.\ SimCLR, at each label budget used in the main paper (1/2/4 labels per class). 1-NN accuracy depends only on the frozen embedding and the labeled set for a given seed, not on \VAST, the propagation graph, or any distillation gate, so it isolates embedding quality from every algorithmic choice downstream of the encoder. Figure~\ref{fig:tinyimagenet_1nn} shows SimCLR beating BYOL by a wide margin at every budget (14.6 vs.\ 8.4 at 1 label/class, 18.2 vs.\ 10.3 at 2 labels/class, 21.6 vs.\ 12.3 at 4 labels/class, roughly $1.7$--$1.8\times$ higher accuracy under SimCLR throughout), corroborating the purity-based argument in the main paper (Section~4.1, ``Embeddings'') with an independent signal pointing to the same conclusion: BYOL is a substantially weaker embedding than SimCLR specifically on TinyImageNet, motivating the switch. The BYOL numbers in both figures are drawn from the BYOL-era TinyImageNet run that preceded the SimCLR switch; because purity and 1-NN accuracy are computed directly from the frozen embedding and do not involve \VAST's training procedure, both remain valid embedding-quality comparisons.

\begin{figure}[tb]
\centering
\includegraphics[width=0.9\linewidth]{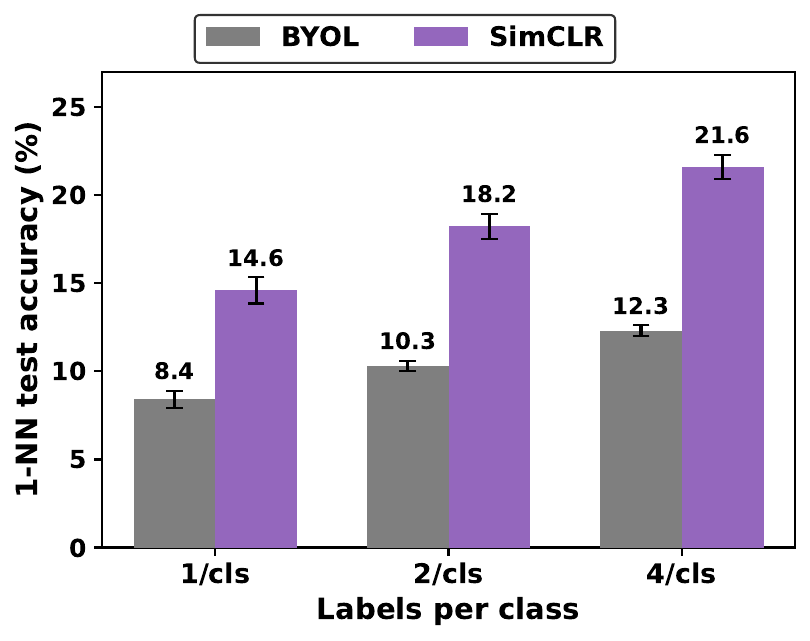}
\caption{TinyImageNet 1-NN test accuracy (mean $\pm$ SEM over 5 seeds) under BYOL vs.\ SimCLR, at 1/2/4 labels per class.}
\label{fig:tinyimagenet_1nn}
\end{figure}


\subsection{Match-Family Collapse Diagnostics}
\label{app:match_collapse}

Section~4.2 of the main paper reports that the *Match-family baselines collapse in the cold-start regime (e.g.\ \FreeMatch at 9.8\% vs.\ \VAST at 39.2\%, CIFAR-100, 100 labels), attributing this to training being dominated by confidently-wrong pseudo-labels once the model overfits the small labeled set. This subsection diagnoses that mechanism directly from each method's own per-iteration training logs (CIFAR-100, 100 labels), rather than inferring it from the final accuracy alone.

We instrument \texttt{util\_ratio}, the fraction of the unlabeled batch whose pseudo-label confidence clears the method's threshold at a given step and is already logged natively by both methods, alongside the unsupervised (pseudo-label) loss. Figure~\ref{fig:match_collapse} shows two distinct failure modes, not one:

\textbf{FreeMatch (poisoning).} \texttt{util\_ratio} saturates near 1.0 within the first few hundred iterations and remains at 0.87--0.998 for the entire $2^{20}{\approx}1048$K-iteration run (mean 0.92), while the unsupervised loss never approaches the near-zero level reached by the supervised loss. Since \FreeMatch's only unsupervised signal is confidence-gated, an open gate combined with a persistently high unsupervised loss indicates the model overfits the 100 labeled points almost immediately, becomes overconfident, and floods training with pseudo-labels that clear the confidence threshold while remaining largely incorrect, for the entire run and not just early on.

\textbf{CoMatch (starvation, partially rescued).} \texttt{util\_ratio} collapses to exactly 0.0 within about 1,300 iterations and stays there for the remainder of the run, the opposite pattern from \FreeMatch. \CoMatch's confidence-gated pseudo-label pathway starves almost immediately. It nonetheless reaches a substantially higher final accuracy (30.7\% vs.\ \FreeMatch's 9.8\%) because its training objective, $\mathcal{L} = \mathcal{L}_{\text{sup}} + \lambda_u \mathcal{L}_{\text{unsup}} + \lambda_c \mathcal{L}_{\text{contrast}}$, includes a second, \emph{ungated} contrastive graph-regularization term built from the full pairwise pseudo-label similarity matrix (no confidence cutoff), which continues to supply a geometry-based training signal even while the confidence-gated pathway is fully starved.

These two mechanisms, poisoning when the entire unsupervised signal is confidence-gated versus partial recovery when an ungated geometric-smoothness signal is also present, are consistent with, and help motivate, \VAST's design: constructing pseudo-label evidence entirely from labeled-set geometry, with no dependence on model confidence at any point, avoids both failure modes.

\begin{figure}[t]
\centering
\includegraphics[width=\linewidth]{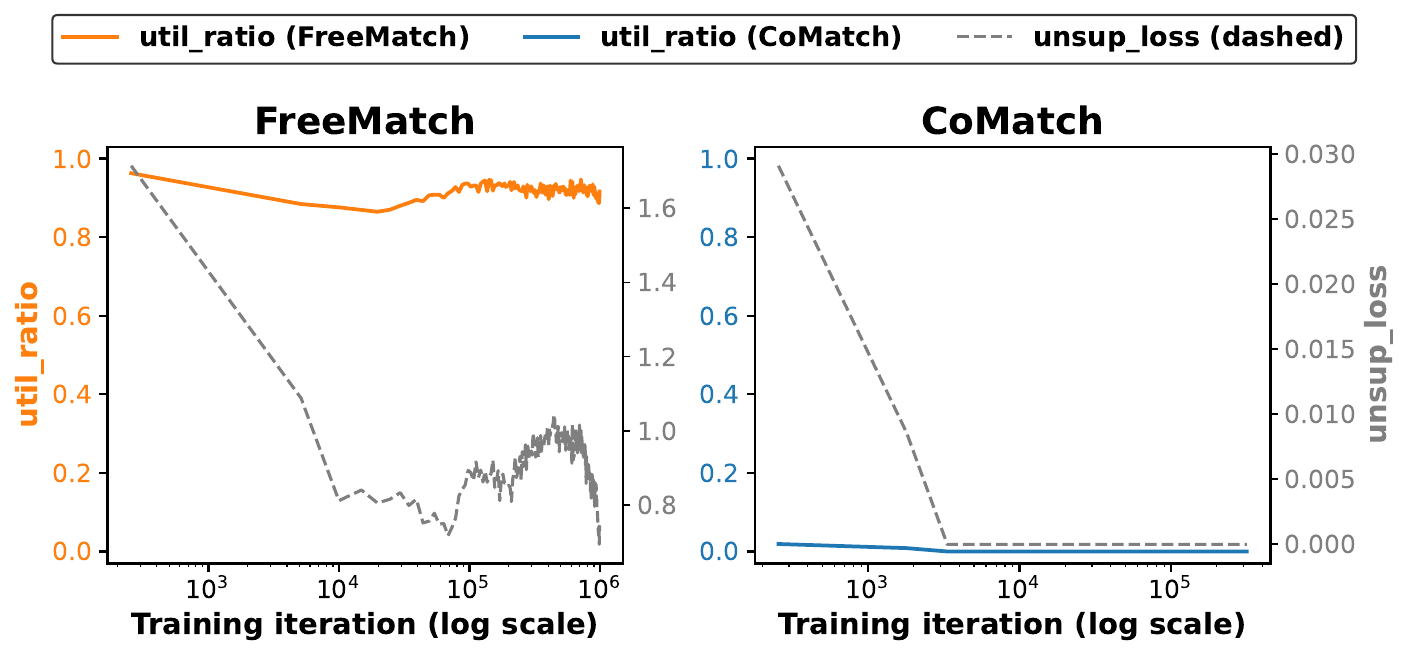}
\caption{Per-iteration \texttt{util\_ratio} (solid, left axis) and unsupervised loss (dashed, right axis, gray) for \FreeMatch and \CoMatch, CIFAR-100, 100 labels. \CoMatch's ungated contrastive graph loss is not shown.}
\label{fig:match_collapse}
\end{figure}


\subsection{Conf-ST: Confidence-Gated Self-Training on Frozen Features}
\label{app:conf_st}

Conf-ST is \VAST\ with the veracity beliefs removed. Soft targets and the admission score are taken from the classifier's own softmax, $p_i = \mathrm{softmax}(f(\Phi(x_i)))$ with gate score $\max_c p_{i,c}$, in place of the normalized veracity vector and its maximum. Everything else is held at \VAST's setting: the same frozen BYOL features, the same classifier and optimizer, the same distillation loss and weight $\lambda$, the same labeled sets per seed, and the same seeds. Targets are recomputed from the previous round's model and detached; the classifier is re-initialized each round. The adaptive variant replicates \FreeMatch's self-adaptive threshold, an EMA of mean confidence with per-class modulation, on these frozen features. Table~\ref{tab:conf_st_sweep} sweeps the gate over five settings, all averaged over the same 5 seeds as \VAST.

\begin{table}[tb]
\centering
\small
\begin{tabular}{@{}lccc@{}}
\toprule
Gate & 100 & 200 & 400 \\
\midrule
$\tau = 0.60$ & $22.3_{\pm8.0}$ & $39.4_{\pm0.5}$ & $47.4_{\pm0.7}$ \\
$\tau = 0.75$ & $10.9_{\pm6.4}$ & $38.4_{\pm0.6}$ & $\mathbf{48.2_{\pm0.7}}$ \\
$\tau = 0.90$ & $11.7_{\pm0.9}$ & $38.1_{\pm0.4}$ & $46.7_{\pm0.8}$ \\
$\tau = 0.95$ & $\mathbf{30.4_{\pm0.3}}$ & $\mathbf{39.9_{\pm1.0}}$ & $47.2_{\pm0.6}$ \\
adaptive      & $27.2_{\pm0.2}$ & $37.1_{\pm0.6}$ & $46.6_{\pm0.7}$ \\
\midrule
MLP \scriptsize{(labeled only)} & $29.4_{\pm0.5}$ & $38.8_{\pm0.5}$ & $47.6_{\pm0.5}$ \\
\VAST & $\mathbf{39.2_{\pm0.9}}$ & $\mathbf{47.8_{\pm0.3}}$ & $\mathbf{54.4_{\pm0.4}}$ \\
\bottomrule
\end{tabular}
\caption{Conf-ST threshold sweep on CIFAR-100 (BYOL), mean $\pm$ SEM over 5 seeds. Bold marks the best gate per budget, the values reported in Table~\ref{tab:track2} of the main paper.}
\label{tab:conf_st_sweep}
\end{table}

Two features of the sweep matter. First, the gate barely helps even at its best: at 100 labels only the strictest setting ($\tau{=}0.95$) exceeds labeled-only training at all, while $\tau{=}0.75$ and $\tau{=}0.90$ fall roughly 18 points below it, since confidence gating on these features is actively harmful over most of its range, not merely weaker than \VAST. Second, the 100-label column is not just low but unstable, with per-seed SEMs up to $8.0$ against $0.6$--$0.8$ at 400 labels: with one label per class the classifier's own confidence is not a reproducible ranking signal, which is precisely the cold-start failure the Veracity Matrix is designed to sidestep. Against \VAST, the best Conf-ST gate per budget is significantly worse at all three budgets (paired $t$-test on shared seeds: $p{=}0.0001$ at 100 labels, $p{=}0.0022$ at 200 labels, $p{=}0.0037$ at 400 labels; \VAST\ wins 5/5 seeds at every budget).


\subsection{Time and Space Complexity}
\label{app:complexity}

\VAST decomposes into a \emph{geometry phase} (executed once per labeled set, prior to classifier training) and a \emph{learning phase} (standard SGD). The geometry phase is amortized over all subsequent training epochs and inference queries; its cost is analyzed below alongside inference, with the learning phase noted only for completeness since its complexity is identical in form to training any classifier of the same architecture.

Let $N=|\mathcal L\cup\mathcal U|$, $d$ the embedding dimension, $k$ the connectivity-selected neighborhood size (Section~\ref{subsec:sparsity} of the main paper), $M$ the number of classes, and $T$ the number of propagation rounds (Section~\ref{subsec:propagation} of the main paper). Table~\ref{tab:complexity} summarizes the cost of each stage.

\begin{table*}[tb]
\centering
\small
\setlength{\tabcolsep}{6pt}
\begin{tabular}{@{}p{0.14\textwidth}p{0.38\textwidth}p{0.28\textwidth}p{0.08\textwidth}@{}}
\toprule
Stage & Time & Space & Frequency \\
\midrule
Graph construction & $O(N^2 d)$ with brute-force pairwise distances; reduced to sub-quadratic in practice by any standard exact or approximate nearest-neighbor index (e.g.\ FAISS, HNSW) without changes to the rest of the algorithm & $O(Nk)$, sparse & once \\
Veracity Matrix seeding & $O\!\left(\sum_{j\in\mathcal L}\deg(x_j)\right)$, averaging to $O(N_L k)$ under the sparse $k$-NN graph & $O(NM)$ for the matrix itself the dominant memory structure for typical $N,M$ (e.g.\ $5{\times}10^6$ floats on CIFAR-100, vs.\ ${\sim}6.5{\times}10^5$ pre-symmetrization directed edges, ${\sim}5{\times}10^5$ after symmetrization) & once \\
Propagation & $O\!\left(M\sum_t |\mathcal R_t| k\right)$ only active receivers are updated each round, and $|\mathcal R_t|$ monotonically shrinks as points cross $\tau_r$ and freeze; $O(TNkM)$ is a loose upper bound & reuses seeding & once \\
Classifier training & $O(E\cdot|\mathcal T_{\text{train}}|\cdot F)$ for $E$ epochs at per-example cost $F$ identical in form to any baseline of the same architecture & model-dependent & every epoch \\
Inference (new test point) & $O(F)$ one forward pass through $f_\theta$ & model-dependent & every query \\
\bottomrule
\end{tabular}
\caption{Asymptotic time and space complexity of \VAST, decomposed by stage. The connectivity criterion selected $k\in\{3,5,13\}$ across the three evaluated datasets.}
\label{tab:complexity}
\end{table*}

\paragraph{Inductive deployment.} Once trained, \VAST classifies a new point with a single forward pass through $f_\theta$, with no dependence on $N$ and no need to touch the graph or the Veracity Matrix again. In contrast, the transductive baselines \Poisson Learning, \Laplace/LP, and \CS must re-solve the propagation system on the enlarged $(N{+}1)$-node graph per query, at cost at least $O(|E|)$ plus solver overhead. \citet{iscen2019label} is inductive like \VAST, training a classifier on LP-derived pseudo-labels, so its test-time cost is likewise a single forward pass (\app~\ref{app:timing}). This is the asymptotic form of the deployment-cost argument in Section~\ref{sec:related_work} of the main paper, and the empirical form is quantified in the timing analysis of Section~\ref{subsec:main_results} of the main paper and \app~\ref{app:timing}.


\subsection{Full Timing Results}
\label{app:timing}

\begin{table*}[tb]
\centering
\small
\setlength{\tabcolsep}{6pt}
\begin{tabular}{@{}lcccccc@{}}
\toprule
 & \multicolumn{3}{c}{CIFAR-100} & \multicolumn{3}{c}{ImageNet100} \\
\cmidrule(lr){2-4} \cmidrule(lr){5-7}
Method & 1/cls & 2/cls & 4/cls & 1/cls & 2/cls & 4/cls \\
\midrule
\Poisson ($k{=}50$)    & $117.7_{\pm5.9}$ & $158.7_{\pm1.2}$ & $203.7_{\pm3.7}$ & $595.7_{\pm0.3}$ & $472.3_{\pm117.1}$$^\dagger$ & $628.3_{\pm2.9}$ \\
\Laplace (LP)           & $270.0_{\pm2.1}$ & $246.0_{\pm4.0}$ & $192.7_{\pm4.9}$ & $287.3_{\pm3.9}$ & $323.3_{\pm34.7}$ & $298.3_{\pm38.4}$ \\
C\&S                   & $158.3_{\pm1.8}$ & $140.3_{\pm7.3}$ & $131.7_{\pm0.9}$ & $223.0_{\pm2.7}$ & $240.7_{\pm15.6}$ & $213.0_{\pm33.5}$ \\
\Iscen & $78.7_{\pm1.2}$ & $77.0_{\pm1.0}$ & $79.3_{\pm0.9}$ & $223.7_{\pm0.9}$ & $260.3_{\pm8.1}$ & $235.0_{\pm16.1}$ \\
\midrule
\textbf{VAST (distill + prop.)} & $\mathbf{18.6_{\pm0.8}}$ & $\mathbf{22.3_{\pm0.1}}$ & $\mathbf{22.5_{\pm0.8}}$ & $\mathbf{142.3_{\pm7.3}}$ & $\mathbf{142.8_{\pm6.3}}$ & $\mathbf{139.8_{\pm2.3}}$ \\
\bottomrule
\end{tabular}
\caption{Track~1 timing: end-to-end time in seconds (training plus test-set prediction) on identical BYOL embeddings, averaged over 3 seeds. Label budgets are per class.}
\label{tab:track1_timing}

{\footnotesize $^\dagger$\Poisson's solver converged ${\sim}2.5\times$ faster on one seed (238s vs.\ 590/589s), reproducibly across reruns: seed-to-seed variance in convergence, not a measurement artifact.}
\end{table*}
Table~\ref{tab:track1_timing} reports the full per-configuration Track-1
end-to-end times summarized in Section~4.2 of the main paper, across both
CIFAR-100 and ImageNet100. Figure~\ref{fig:track2_timing} of the main paper reports the Track-2
full-pipeline cost comparison summarized there.

All times include test-set prediction, which the two families account for differently: the transductive methods (\Poisson, \Laplace, \CS) place the 10K test nodes in the graph and obtain test predictions as a byproduct of the solve, while for the inductive methods (\Iscen, \VAST) the test-set forward pass adds a negligible fraction of a second.
Beyond raw speed, that distinction matters for deployment: classifying a new point costs an inductive method one forward pass, while a transductive method must rebuild the graph with the new points included and re-solve the full system, a cost that scales with dataset size and not with the number of new points.

\VAST's total cost in the full-pipeline (Track~2) setting is one-time BYOL pretraining (CIFAR-100: 12.8h; ImageNet100: 39.9h), amortized across every seed and label budget, plus a negligible per-run cost (${\sim}$19s / ${\sim}$142s at 1 label per class).
\FreeMatch has no amortizable stage and pays its full end-to-end cost on every run: $261.1_{\pm30.5}$h on CIFAR-100 (2 seeds, ${\sim}2^{20}$ iterations) and 159h on ImageNet100 (single seed, at the 472K-iteration checkpoint at which its accuracy is reported a lower bound on the full schedule).
Even including pretraining, \VAST is ${\sim}20\times$ cheaper than a single \FreeMatch run on CIFAR-100 and ${\sim}4\times$ cheaper on ImageNet100, and the gap widens with every additional seed or budget evaluated.


\subsection{Cross-Dataset Propagation Ablation}
\label{app:prop_ablation_full}

Table~\ref{tab:prop_ablation_full} extends the CIFAR-100 propagation ablation
of Section~4.3 of the main paper to ImageNet100. Both
distill-only variants use a constant distillation threshold $\tau$, per
Section~4.3 of the main paper, and additionally a per-variant $\alpha_0$
re-tuned for each variant on ImageNet100, a detail specific to this
extension and not stated in the main paper's CIFAR-100 ablation; the
full-method row is repeated from the main paper's cross-dataset table for
reference.

\begin{table*}[tb]
\centering
\small
\setlength{\tabcolsep}{6pt}
\begin{tabular}{@{}lcccccc@{}}
\toprule
& \multicolumn{3}{c}{CIFAR-100} & \multicolumn{3}{c}{ImageNet100} \\
\cmidrule(lr){2-4} \cmidrule(lr){5-7}
Configuration & 1/cls & 2/cls & 4/cls & 1/cls & 2/cls & 4/cls \\
\midrule
\textbf{VAST (distill + prop.)} & $\mathbf{39.2_{\pm0.9}}$ & $\mathbf{47.8_{\pm0.3}}$ & $\mathbf{54.4_{\pm0.4}}$ & $\mathbf{34.9_{\pm1.1}}$ & $\mathbf{43.1_{\pm0.9}}$ & $\mathbf{49.3_{\pm0.2}}$ \\
Distill only (matched kernel) & $30.5_{\pm0.7}$ & $38.9_{\pm0.6}$ & $47.3_{\pm0.7}$ & $25.0_{\pm0.9}$ & $34.5_{\pm0.7}$ & $40.7_{\pm0.6}$ \\
Distill only (local RBF, best-effort) & $34.9_{\pm1.1}$ & $42.0_{\pm0.9}$ & $48.8_{\pm0.3}$ & $27.1_{\pm0.9}$ & $36.0_{\pm1.2}$ & $42.9_{\pm0.2}$ \\
\bottomrule
\end{tabular}
\caption{Propagation ablation on CIFAR-100 and ImageNet100: test accuracy (\%) by configuration and label budget (labels per class).}
\label{tab:prop_ablation_full}
\end{table*}

\subsection{Degree-Normalization Ablation: Significance}
\label{app:degree_ablation_significance}

Per-budget statistics for Table~\ref{tab:degree_ablation} of the main paper, paired $t$-test on shared seeds: 1 label/class, 39.2 vs.\ 37.8, $p{=}0.02$; 2 labels/class, 47.8 vs.\ 46.8, $p{=}0.06$; 4 labels/class, 54.4 vs.\ 53.7, $p{=}0.04$. \VAST\ wins on 5/5 seeds at every budget.

\subsection{Distillation Confidence Weighting Ablation}
\label{app:distill_weight_ablation}

Section~3.5 of the main paper weights the distillation loss per-sample by each point's own peakedness $c_i$. Table~\ref{tab:distill_weight_ablation} compares this against uniform weighting, i.e.\ a constant weight of 1 for every admitted point, holding the gate, propagation, and all other hyperparameters fixed. On CIFAR-100, the two variants are statistically indistinguishable at every budget: the largest gap (0.26 points at 200 labels) is well within one seed-level SEM. On TinyImageNet (SimCLR embedding, 5 seeds), uniform weighting is likewise indistinguishable from the reported per-sample result at every budget (largest gap 0.41 points at 800 labels, well within one seed-level SD). The presence of confidence weighting itself does not measurably change \VAST's accuracy on either dataset. We adopt per-sample weighting because it is the simpler, more standard confidence-weighted-distillation formulation and each point's own confidence is already computed for the gate (Section~3.5); the ablations show this choice is not necessary for the reported gains, only that it is harmless.

\begin{table}[tb]
\centering
\small
\setlength{\tabcolsep}{3pt}
\begin{tabular}{@{}lccc@{}}
\toprule
\multicolumn{4}{c}{CIFAR-100} \\
\cmidrule(lr){1-4}
Weighting & 1/cls & 2/cls & 4/cls \\
\midrule
Per-sample $c_i$ (reported) & $39.2_{\pm0.9}$ & $47.8_{\pm0.3}$ & $54.4_{\pm0.4}$ \\
Uniform ($w_i = 1$) & $39.06_{\pm0.76}$ & $47.54_{\pm0.16}$ & $54.23_{\pm0.26}$ \\
\midrule
\multicolumn{4}{c}{TinyImageNet} \\
\cmidrule(lr){1-4}
Weighting & 200 lbl & 400 lbl & 800 lbl \\
\midrule
Per-sample $c_i$ (reported) & $17.71_{\pm1.26}$ & $22.83_{\pm1.40}$ & $27.55_{\pm1.61}$ \\
Uniform ($w_i = 1$) & $17.58_{\pm1.19}$ & $22.88_{\pm1.35}$ & $27.96_{\pm0.95}$ \\
\bottomrule
\end{tabular}
\caption{Distillation weighting ablation, mean $\pm$ dispersion over 5 seeds (SEM for CIFAR-100; sample SD for TinyImageNet, \app~\ref{app:implementation}). CIFAR-100 uses BYOL; TinyImageNet uses SimCLR.}
\label{tab:distill_weight_ablation}
\end{table}


\subsection{Peakedness Function Choice}
\label{app:peakedness}

The peakedness score $c_k = \max_c \tilde V_{k,c}$ (Section~3.2 of the main paper) is the only summary of a point's belief vector that \VAST\ uses, setting the propagation sender and receiver sets (Section~3.4), the distillation gate, and the per-sample distillation weight (Section~3.5). Writing $\tilde V_{k,(1)} \geq \tilde V_{k,(2)}$ for the two largest entries of $\tilde{\mathbf{V}}_k$, we compare Max, $\tilde V_{k,(1)}$; Margin, $\tilde V_{k,(1)} - \tilde V_{k,(2)}$; and Entropy, $1 - H(\tilde{\mathbf{V}}_k)/\log M$, normalized and inverted so that larger values indicate greater peakedness in all three cases. Each score is substituted into an otherwise identical pipeline (Table~\ref{tab:peakedness_ablation}).

Max and Margin are indistinguishable, differing by at most $0.31$ points, below one seed-level SEM at every budget and without a consistent sign; both are determined primarily by the largest entry of the belief vector and induce nearly identical rankings. Entropy trails Max at every budget, by $1.68$, $1.61$, and $1.46$ points, exceeding twice the combined SEM at 2 labels per class. Under Entropy, \VAST\ still leads \Poisson\ Learning at 2 and 4 labels per class, but falls slightly below it at 1 ($37.52$ vs.\ $38.6$, cf.\ Table~\ref{tab:crossdataset} of the main paper), the one budget where the Max-based lead was already not significant (\app~\ref{app:track1_significance}).

The three scores span at most $1.9$ points, against \VAST's $2.7$- and $4.3$-point margins over \Poisson\ Learning at 2 and 4 labels per class. This spread, rather than exact interchangeability, is the sense in which Section~3.2 of the main paper describes performance as robust to this choice. We attribute Entropy's deficit to gate calibration rather than to information content: \VAST's gate constants are stated on the scale of Max, since the sender rule $c_i > 1/M$ and the floor $\tau_c \geq 1/M$ both correspond to the score that a \emph{uniform} belief receives under Max, so a score of a different shape shifts the effective threshold in addition to the ranking. Isolating the two effects would require re-deriving the gate for each score, which we do not attempt here.

We therefore adopt Max: it is the maximum entry of the posterior mean under the Dirichlet interpretation (\app~\ref{app:bayesian_justification}), the gate constants are natural on its scale, and it matches or exceeds both alternatives at every budget. The unnormalized-margin variant noted in Section~3.2 of the main paper is reported from that earlier exploration and was not re-run under the current pipeline: it is not a function on the simplex, as raw row sums vary with the kernel mass each point receives from $\mathcal{L}$, so it conflates the amount of evidence available at a point with the unambiguity of that evidence, and its unbounded range would require re-deriving the gate rather than substituting the score directly.

\begin{table}[tb]
\centering
\small
\setlength{\tabcolsep}{3pt}
\begin{tabular}{@{}lccc@{}}
\toprule
\multicolumn{4}{c}{CIFAR-100 (BYOL)} \\
\cmidrule(lr){1-4}
Peakedness function & 1/cls & 2/cls & 4/cls \\
\midrule
Max, $\tilde V_{k,(1)}$ (reported) & $39.2_{\pm0.9}$ & $47.8_{\pm0.3}$ & $54.4_{\pm0.4}$ \\
Margin, $\tilde V_{k,(1)}-\tilde V_{k,(2)}$ & $39.41_{\pm0.93}$ & $47.49_{\pm0.21}$ & $54.29_{\pm0.44}$ \\
Entropy, $1-H(\tilde{\mathbf{V}}_k)/\log M$ & $37.52_{\pm0.78}$ & $46.19_{\pm0.27}$ & $52.94_{\pm0.52}$ \\
\midrule
\Poisson\ (reference) & $38.6_{\pm1.0}$ & $45.1_{\pm0.7}$ & $50.1_{\pm0.4}$ \\
\bottomrule
\end{tabular}
\caption{Peakedness function ablation on CIFAR-100, mean $\pm$ SEM over 5 seeds, current pipeline. Max and Margin are within one SEM of each other at every budget and are therefore left unbolded; Entropy trails both at every budget. \Poisson\ Learning is repeated from Table~\ref{tab:crossdataset} of the main paper as a reference point.}
\label{tab:peakedness_ablation}
\end{table}


\subsection{Additional Embedding Results}
\label{app:embeddings}

\paragraph{Barlow Twins and SimCLR.}
To test whether \VAST's algorithmic contribution generalizes beyond BYOL, we
evaluate \VAST and the strongest label-propagation baselines (\Poisson, \PoissonMBO, \Laplace) on Barlow Twins and SimCLR embeddings, alongside the BYOL
results from the main paper (Table~\ref{tab:embeddings}).
Barlow Twins uses no momentum target or stop-gradient, so a matching result
rules out \VAST's gains being an artifact of BYOL's particular training
dynamics; \VAST outperforms every graph-diffusion baseline evaluated here on Barlow Twins at every budget while
trailing its own BYOL results only slightly.
SimCLR instead serves as a lower-quality-embedding stress test, probing how
\VAST degrades when the smoothness assumption (Assumption~1 of the main
paper) holds less well, rather than as a claim of generality: \VAST degrades
gracefully: accuracy drops with embedding quality, as the assumption
predicts, while still matching or exceeding \Poisson Learning at every
budget on the identical SimCLR features.

\begin{table}[tb]
\centering
\small
\setlength{\tabcolsep}{4pt}
\begin{tabular}{@{}llccc@{}}
\toprule
Embed. & Method & 100 & 200 & 400 \\
\midrule
BYOL
  & \VAST          & $\mathbf{39.2_{\pm0.9}}$ & $\mathbf{47.8_{\pm0.3}}$ & $\mathbf{54.4_{\pm0.4}}$ \\
\midrule
\multirow{4}{*}{Barlow Tw.}
  & \VAST          & $\mathbf{37.3_{\pm0.5}}$ & $\mathbf{44.7_{\pm0.4}}$ & $\mathbf{51.8_{\pm0.4}}$ \\
  & \Poisson       & $36.37_{\pm0.8}$ & $43.2_{\pm0.5}$ & $48.49_{\pm0.1}$ \\
  & \PoissonMBO   & $34.8_{\pm0.8}$ & $43.4_{\pm0.6}$ & $49.6_{\pm0.4}$ \\
  & \Laplace       & $24.2_{\pm1.1}$ & $37.4_{\pm0.5}$ & $45.9_{\pm0.2}$ \\
\midrule
\multirow{4}{*}{SimCLR}
  & \VAST          & $\mathbf{28.9_{\pm0.4}}$ & $\mathbf{34.9_{\pm0.3}}$ & $\mathbf{41.3_{\pm0.4}}$ \\
  & \Poisson       & $26.83_{\pm0.5}$ & $31.4_{\pm0.4}$ & $36.17_{\pm0.5}$ \\
  & \PoissonMBO   & $26.3_{\pm0.7}$ & $30.3_{\pm0.7}$ & $37.8_{\pm0.7}$ \\
  & \Laplace ($k{=}20$) & $17.2_{\pm0.7}$ & $26.0_{\pm0.4}$ & $33.8_{\pm0.4}$ \\
\bottomrule
\end{tabular}
\caption{Test accuracy (\%) on CIFAR-100 with BYOL, Barlow Twins, and SimCLR embeddings, random label selection.}
\label{tab:embeddings}

{\footnotesize All Barlow Twins baselines use $k{=}50$; SimCLR \Laplace uses $k{=}20$ as noted.}
\end{table}

\paragraph{VAE2: a baseline-attribution check.}
Every embedding compared so far is used by every Track~1 method that column evaluates, so within-column differences are attributable to the algorithm rather than the representation (Section~\ref{subsec:main_results} of the main paper). VAE2, the variational-autoencoder embedding distributed with the \Poisson\ Learning library~\citep{calder2020poisson} itself (i.e.\ its own authors' recommended default), is used to ask a different, narrower question: does \Poisson's strength in Track~1 come from the solver, or from the BYOL embedding it happens to be evaluated on there? \Poisson\ collapses to $4.5_{\pm0.2}$\% at 100 labels on VAE2 (CIFAR-100), versus $38.6_{\pm1.0}$\% on BYOL, a $34$-point drop on the identical solver, dataset, and labeled sets, with only the embedding changed. Table~\ref{tab:vae2} reports the full three-budget comparison.

\begin{table}[t]
\centering
\small
\begin{tabular}{@{}lccc@{}}
\toprule
Embedding & 100 & 200 & 400 \\
\midrule
BYOL  & $38.6_{\pm1.0}$ & $45.1_{\pm0.7}$ & $50.1_{\pm0.4}$ \\
VAE2  & $4.5_{\pm0.2}$ & $5.4_{\pm0.3}$ & $7.1_{\pm0.2}$ \\
\bottomrule
\end{tabular}
\caption{\Poisson\ Learning test accuracy (\%) on BYOL vs.\ VAE2 embeddings, CIFAR-100. Same solver, dataset, and labeled sets; only the embedding differs. VAE2 is the \Poisson\ Learning library's own recommended default, so this is not an adversarially chosen embedding.}
\label{tab:vae2}
\end{table}

This is why the VAE2 result is a check on Track~1's design rather than a fourth full comparison: it confirms that Track~1's within-column comparability, every method sharing identical frozen features, is doing real work, since pulling that shared embedding out from under a graph-based baseline is sufficient to collapse it. \VAST, \CS, and 1-NN are not run on VAE2, since the check only requires isolating one baseline's dependence on embedding quality, not a full re-evaluation of every method on a fourth embedding.

\paragraph{DINOv2 (foundation-model embedding).}
Since the SSL encoders evaluated above (BYOL, Barlow Twins, SimCLR) are all
of a broadly similar generation and training scale, we additionally evaluate
\VAST\ against the strongest Track~1 graph baselines (\Poisson, \CS, \Laplace,
\Iscen) on DINOv2~\citep{oquab2024dinov2} (ViT-B/14)
features, a foundation-model embedding with substantially larger scale and
pretraining data than the SSL encoders used elsewhere in the paper, to test
whether \VAST's algorithmic contribution is specific to the SSL-scale regime
or persists on modern foundation representations. All methods share identical
DINOv2 features per seed, following the Track~1 protocol
(Section~\ref{subsec:main_results} of the main paper); results in
Table~\ref{tab:dinov2}, mean~$\pm$~SEM over 5 seeds, with paired
$t$-tests on shared labeled sets against \VAST. The 1-NN control on DINOv2 is
reported in Table~\ref{tab:embedding_sweep} of the main paper ($58.4$ at 1
label per class, $72.56$ at 4), which \VAST\ leads by $+11.7$ points at one
label per class; we compare here against the graph baselines, over which the
cold-start margin is the substantive question.

\begin{table}[tb]
\centering
\small
\setlength{\tabcolsep}{5pt}
\begin{tabular}{@{}lccc@{}}
\toprule
Method & 100 lbl & 200 lbl & 400 lbl \\
\midrule
\Laplace                & $57.12_{\pm0.77}$ & $70.16_{\pm0.55}$ & $77.77_{\pm0.37}$ \\
\Iscen & $46.52_{\pm6.90}$$^\dagger$ & $71.20_{\pm1.37}$ & $77.80_{\pm0.73}$ \\
\CS     & $67.60_{\pm0.75}$ & $\mathbf{77.74_{\pm0.13}}$ & $\mathbf{82.07_{\pm0.16}}$ \\
\Poisson                & $\mathbf{69.81_{\pm0.89}}$ & $\mathbf{77.41_{\pm0.13}}$ & $79.73_{\pm0.34}$ \\
\midrule
\textbf{VAST (distill + prop.)} & $\mathbf{70.07_{\pm0.68}}$ & $\mathbf{78.28_{\pm0.79}}$ & $\mathbf{82.83_{\pm0.54}}$ \\
\bottomrule
\end{tabular}
\caption{Track~1 comparison on DINOv2 (ViT-B/14) features, CIFAR-100; mean $\pm$~SEM over 5 seeds with identical labeled sets per seed. Among the baselines, bold marks the strongest baseline in each column together with any baseline not significantly distinguishable from \VAST\ ($p\geq0.05$, paired $t$-test); \Laplace\ and \Iscen\ are never bolded, since no significance test was run for those two comparisons. Significance values are discussed in the text.}
\label{tab:dinov2}

{\footnotesize $^\dagger$Bimodal across seeds; see text.}
\end{table}

\Iscen's 100-label entry warrants a note: one seed collapses to $20.79\%$ while the other four sit in the $47$--$60\%$ range, giving a mean of $52.96\%$ (sample SD $6.38$ over the remaining four seeds, not a SEM and not comparable to the table's $\pm6.90$ SEM over all five) once the outlier is dropped, still far behind \VAST's $70.07\%$ but no longer bimodal. We read this as a genuine LP-pseudo-label convergence failure at extreme low budget on DINOv2 features rather than a measurement artifact, though we have not investigated it further.

On a foundation-scale embedding \VAST\ leads the diffusion-based \Poisson\
baseline significantly at the highest budget ($+3.1$ pts, $p{=}0.009$) and is
indistinguishable from it at the two lower ones, while leading \CS\
significantly at the lowest budget ($+2.5$ pts, $p{=}0.045$, 4/5 seeds)
before converging to parity at 200 and 400 labels ($p{=}0.59$, $p{=}0.28$);
\VAST's raw margin over \Laplace and \Iscen is positive at every budget (Table~\ref{tab:dinov2}), though no significance test was run for those two comparisons.
The direction of that convergence is the substantive point. \VAST's margin
over \CS, its closest architectural counterpart
(Section~\ref{sec:related_work} of the main paper), is widest where labels are
scarcest and closes as the budget grows, the pattern the cold-start account
predicts, observed here on the strongest representation we evaluate.
The DINOv2 result thus extends the central Track~1 claim, that \VAST\ leads
or matches the strongest transductive baselines on identical embeddings,
from the SSL-scale regime to a foundation-model embedding, while remaining
explicit about where the ties lie (Table~\ref{tab:dinov2}).

\ifshowmain\else
  \bibliography{example_paper}
\fi

\fi

\end{document}